\documentclass[11pt, a4paper, logo, copyright, nonumbering]{xiaomi}

\usepackage[numbers, sort&compress, square]{natbib}
\usepackage{dblfloatfix}
\usepackage[normalem]{ulem}
\usepackage{caption}
\usepackage{dramatist}
\usepackage{xspace}
\usepackage{pifont} 
\usepackage{multirow}
\usepackage{tcolorbox}
\usepackage{xltabular}
\usepackage{longtable}
\usepackage{hyperref}
\usepackage{wrapfig}

\usepackage{amsfonts}
\usepackage{amsmath}
\usepackage{amssymb}
\usepackage{lineno}
\usepackage{multirow}
\usepackage{adjustbox}

\usepackage[bottom]{footmisc}

\usepackage{CJKutf8}
\usepackage{setspace}
\usepackage{makecell}
\usepackage{graphicx}
\usepackage{subcaption}
\usepackage{multicol} 

\usepackage[utf8]{inputenc} 
\usepackage[T1]{fontenc}    
\usepackage{hyperref}
\usepackage{cleveref}
\usepackage{url}            
\usepackage{booktabs}       
\usepackage{amsfonts}       
\usepackage{nicefrac}       
\usepackage{microtype}      
\usepackage{xcolor}         
\usepackage{colortbl}
\usepackage{tcolorbox}
\usepackage{xspace}
\definecolor{BrickRed}{rgb}{.72,0,0}
\definecolor{darkgreen}{rgb}{0.0, 0.5, 0.0}
\definecolor{ForestGreen}{RGB}{34,139,34}
\definecolor{LakeBlue}{RGB}{0,61,153}
\definecolor{MiOrange}{RGB}{255,225,204}
\definecolor{Hex}{RGB}{225,213,231}

\hypersetup{
    colorlinks=true,
    allcolors=LakeBlue
}

\title{\centering PixelWizard: Towards Efficient High-Fidelity Video Generation at Ultra-Large Spatial Resolution}

\titlerunning{PixelWizard}

\author{%
\parbox{\textwidth}{\centering
Wenxue Li$^{1,2*}$, Jingjing Ren$^{1*}$, Peng Zhang$^{2*}$, Tian Ye$^{1}$, Daiguo Zhou$^{2}$, Jian Luan$^{2}$, Xinhu Zheng$^{1}$, Gang Hua$^{4}$, Lei Zhu$^{1,3\dagger}$
}}
\institute{\small
$^1$The Hong Kong University of Science and Technology (Guangzhou), \\ \quad
$^2$MiLM Plus, Xiaomi Inc, \\ \quad
$^3$The Hong Kong University of Science and Technology,\quad
$^4$Amazon
}
\contribution[*]{Equal Contribution}
\contribution[\dagger]{Corresponding Author}
\checkdata[Project Page]{\url{https://wxliii.github.io/pixelwizard/}}

\begin{document}

\begin{abstract}
Scaling video generation to ultra-high spatial resolutions remains challenging due to both optimization difficulty and prohibitive inference cost. On dense high-resolution token grids, global spatiotemporal structures become difficult to optimize jointly with fine-grained textures, while the substantially increased sequence length makes iterative denoising prohibitively expensive. 
To address this, we propose PixelWizard, a framework that hierarchically decouples global structure modeling from fine-grained detail synthesis. 
PixelWizard first constructs a compact spatial-temporal anchor that captures global motion and structural layouts in a semantically dense latent space. The anchor is then incorporated into the high-resolution DiT through a dynamic Anchor-Guided Injector, allowing high-resolution generation to focus on fine-grained detail synthesis under explicit structural guidance. 
Building on this structurally constrained generation process, we further introduce Noise-Span Aligned Shortcut Training for efficient few-step inference. Under a limited denoising step budget, accurate large-span predictions become particularly important. We introduce Exponential Index-Biased Sampling to increase
the sampling frequency of large-span transitions and Adaptive Noise-Span Calibration to increase their relative contribution to the consistency objective.
Together, these designs enable stable few-step high-resolution synthesis without teacher–student distillation. 
Extensive experiments on 2K and 4K video generation demonstrate strong visual quality and over 10× sampling acceleration compared with direct high-resolution extrapolation of the base model.
\end{abstract}

\maketitle

\begin{figure}[hbt!]
    \centering
    \includegraphics[width=\textwidth]{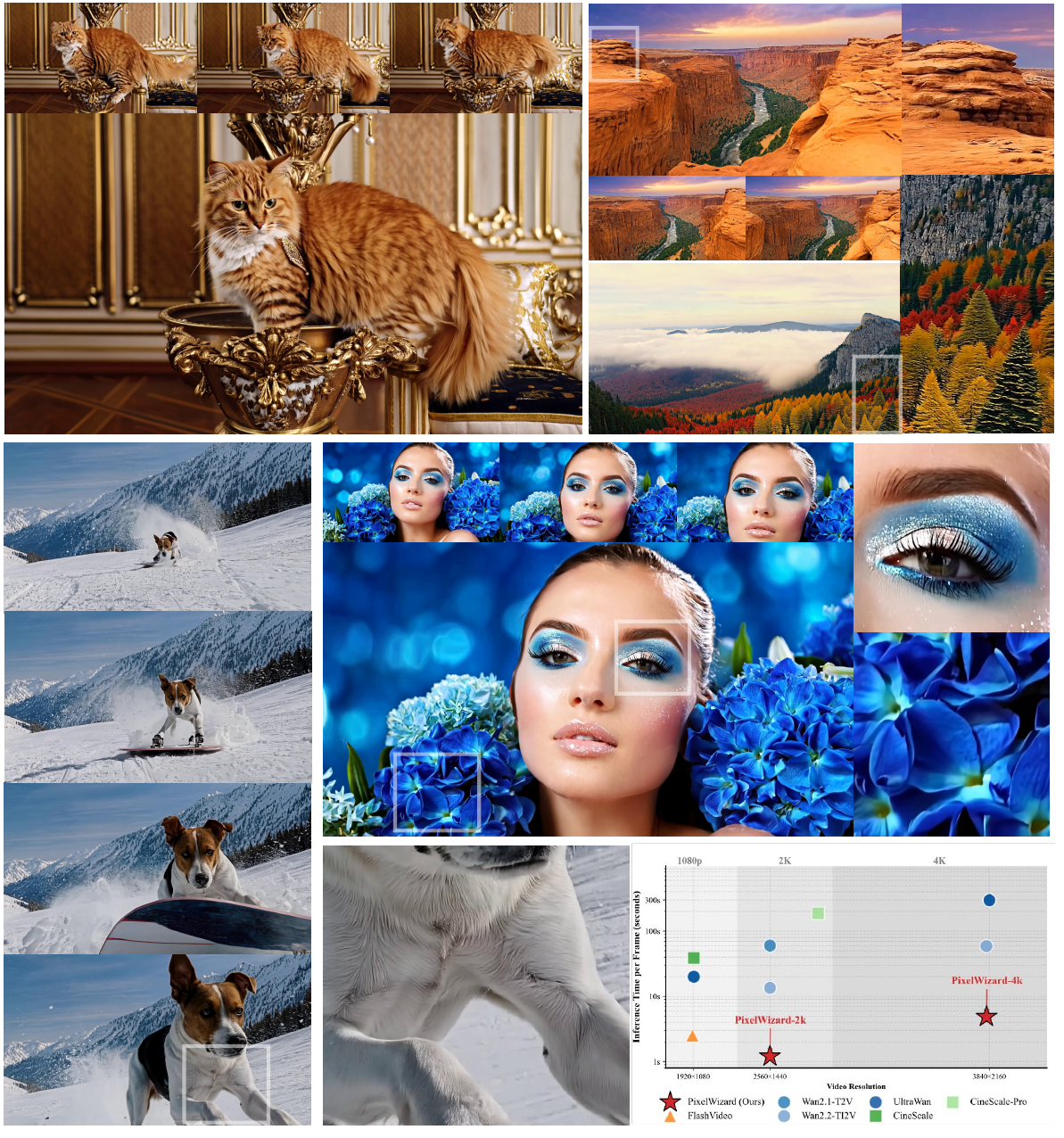}
    \caption{PixelWizard enables high-resolution video synthesis with coherent global structures and fine-grained local details, as revealed by the zoomed-in regions, while maintaining a clear inference speed advantage.}
    
    \label{fig:teaser}
\end{figure}

\section{Introduction}
\label{introduction}

The past few years have ushered in a breakthrough era for video generation. Empowered by powerful diffusion backbones~\cite{DiT} and unprecedented training scales, modern video generative models~\cite{kling25turbo,veo31,sora2,Hunyuanvideo_1.5,LTX-2,Waver_Arxiv25,Ltx-video} are now capable of synthesizing photorealistic, dynamic, and semantically consistent videos directly from text prompts.
However, as expectations for visual fidelity rise, scaling generation to higher resolutions (e.g., 2K–4K) remains a formidable frontier.

To approach this frontier, the community has explored various paradigms, ranging from direct training on high-resolution data~\cite{ultravideo,ultragen} and training-free attention manipulations~\cite{FreeSwim_arxiv2511,FreeScale_CVPR25}, to cascaded generation pipelines~\cite{FlashVideo, Turbo2k_ICCV25}.
While these methods mark significant progress, simply applying them to ultra-high resolutions exposes fundamental limitations in optimization stability and computational efficiency.
This pursuit introduces challenges that can be broadly summarized as three aspects:

\textbf{\textit{(1) Optimization Difficulty: The Semantic Density Dilemma.}}
Modeling high-resolution (HR) videos introduces a semantic sparsity issue. As spatial resolution increases, the semantic information per token becomes increasingly diluted.
Consequently, optimization gradients are dominated by local appearance cues, making it ineffective to jointly optimize global spatiotemporal structure and fine-grained textures.
This structural-textural conflict often leads to artifacts, such as distorted object shapes and repetitive patterns, where the model struggles to maintain global coherence.
\textbf{\textit{(2) Prohibitive Training Overhead.}}
The optimization difficulty directly exacerbates training inefficiency.
Learning reliable long-range spatiotemporal dependencies at scale requires substantially more training iterations, resulting in quadratic growth in computation and memory costs.
Furthermore, the scarcity and high acquisition cost of high-quality 4K video data further constrain large-scale training.
\textbf{\textit{(3) Inference Inefficiency \& The Memory Wall.}}
HR generation incurs prohibitive inference latency as the number of spatiotemporal tokens grows rapidly.
Although distillation can reduce inference cost by shifting part of the burden to training, it introduces substantial hardware demands: at 2K/4K resolutions, HR activations and teacher–student pipelines create a ``memory wall’’ that limits scalability on standard hardware.

These challenges indicate that jointly modeling global structure and fine-grained textures within a single HR stage is fundamentally inefficient.
Motivated by this observation, we propose PixelWizard, a framework that explicitly disentangles structural planning from HR texture synthesis, providing a unified solution to these bottlenecks.
First, Spatial-Temporal Anchor Modeling generates global motion and structural layouts in a compact latent representation, where semantic information is concentrated and long-range dependencies are easier to model. Second, Anchor-Guided High-Resolution Synthesis uses this anchor as an explicit structural prior, allowing the high-resolution model to devote its capacity primarily to fine-grained texture synthesis.
Meanwhile, high-resolution generation remains computationally expensive due to iterative denoising. We therefore introduce Noise-Span Aligned Shortcut Training, which enables few-step inference by learning denoising transitions conditioned on the step size, without memory-intensive teacher–student distillation. 
Under the shifted schedules used for high-resolution generation, few-step inference includes relatively large noise-span transitions.
Uniform sampling over candidate scales does not explicitly prioritize the large-span predictions required in this setting.
Our Exponential Index-Biased Sampling increases exposure to larger spans, while Adaptive Noise-Span Calibration assigns higher relative weights to their consistency losses.
Together, these designs place greater training emphasis on the large-span predictions needed for efficient few-step high-resolution generation.

Our main contributions are summarized as follows:
\begin{itemize}
    \item We propose PixelWizard, a hierarchical framework that separates global spatiotemporal planning from high-resolution texture synthesis. A compact spatial-temporal anchor guides the high-resolution DiT through a dynamic Anchor-Guided Injector, improving structural coherence while easing the modeling of long-range dependencies.
    \item We introduce Noise-Span Aligned Shortcut Training for distillation-free few-step inference. Exponential Index-Biased Sampling and Adaptive Noise-Span Calibration jointly emphasize large noise-span transitions through biased scale sampling and span-dependent consistency weighting.
    \item Extensive experiments on 2K and 4K video generation demonstrate strong generation quality and efficient inference. PixelWizard accelerates sampling by over 10× relative to direct high-resolution extrapolation of the base model, while supporting 4K generation on a single H200 GPU.
\end{itemize}

\section{Related Work}

\subsection{Video Generation}
Early diffusion-based video models extended image diffusion to spatiotemporal generation~\cite{imagenvideo,Make-a-video}, while subsequent works improved scalability through transformer architectures and large-scale video training~\cite{Stable_video_diffusion,opensora-plan,cogvideox}.
Recent advances in diffusion models and diffusion transformers (DiTs)~\cite{DiT} have substantially improved the scalability and generation quality of visual generative models, driving rapid progress in video generation. 
On the closed-source side, commercial systems such as Seedance~\cite{seedance2_0}, Kling~\cite{kling25turbo}, Veo~\cite{veo31}, and Sora~\cite{sora2} have demonstrated impressive visual fidelity, temporal coherence, and complex motion generation. In parallel, open-source models, including HunyuanVideo~\cite{Hunyuanvideo_1.5}, LTX-Video~\cite{Ltx-video}, and Wan~\cite{wan}, have advanced accessible video generation through scalable architectures and large-scale training, providing strong foundations for subsequent research and adaptation.
Despite the rapid progress of both closed- and open-source systems, scaling video generation to ultra-high spatial resolutions remains challenging, as the rapidly growing spatiotemporal token sequence substantially increases both modeling difficulty and computational cost.

\subsection{Video generation at Large Spatial Resolution}
Recent studies have made progress in extending video generation models to higher spatial resolutions and the efforts span several paradigms:
\textit{(i)} Training-free approaches~\cite{FreeSwim_arxiv2511,FreeScale_CVPR25,CineScale_arxiv2508,scalecrafter} avoid tuning overhead, but they often suffer from object repetition, distorted structures and artifacts that undermine visual stability.
Moreover, patch-wise or iterative refinement introduces substantial inference latency, limiting practical efficiency.
\textit{(ii)}
From a training perspective, a straightforward approach is to directly train generative models on high-resolution data~\cite{ultravideo,ultragen,URAE_ICML2025,ultraflux}.
However, the explosive growth of spatiotemporal tokens and reduced semantic density severely hinder optimization, requiring large datasets and slow convergence for long sequences.
\textit{(iii)}
Post-hoc Super-Resolution (SR) methods~\cite{STAR,SeedVR,Seedvr2_arxiv2025,DOVE_Nips25,Flashvsr_Arxiv25,SimpleGVR_Arxiv2506,Vivid-VR} rely heavily on low-resolution inputs, which limits their ability to introduce new high-frequency details when scaling to 2K or beyond.
More recently, growing efforts target high-resolution generation. 
However, existing solutions face efficiency and scalability trade-offs: while UltraGen~\cite{ultragen} and Turbo2k~\cite{Turbo2k_ICCV25} suffer from prohibitive inference latency, methods like FlashVideo~\cite{FlashVideo} and HiStream~\cite{HiStream} fail to scale effectively to ultra-large resolutions.

\subsection{Efficient Video Generation}
Existing efforts on efficient video generation primarily target inference-time acceleration.
Some approaches reduce the number of sampling steps via distillation-based techniques~\cite{DMD,DCM,osv,self-forcing,magicdistillation} or inference-time caching strategies~\cite{teacache,magcache,easycache},
while others aim to lower per-step computation through sparse attention~\cite{SVG,SVG2,radialattention} or latent-space compression~\cite{DC-VideoGen}.
For high-resolution generation, although local attention has been explored~\cite{ultragen}, inference speed remains constrained by the large number of sampling steps, making step reduction the most effective avenue for accelerating generation.
In practice, distillation requires running a heavy teacher–student pipeline in parallel, which introduces prohibitive memory overhead on top of already expensive high-resolution activations and attention, making it infeasible for scalable HR video generation and motivating distillation-free few-step inference.

\section{Method}

\subsection{Preliminaries}
\subsubsection{Flow Matching for Video Generation}

The Rectified Flow~\cite{RectifiedFlow} is utilized to learn a deterministic transport from a standard Gaussian distribution $\pi_0 = \mathcal{N}(0, I)$ to the data distribution $\pi_1 = p_{\text{data}}$.
Let $x_0 \sim \pi_0$ and $x_1 \sim \pi_1$.
Rectified Flow defines a linear probability path $x_t$ for $t \in [0,1]$ as
\begin{equation}
x_t = (1-t)x_0 + t x_1,
\end{equation}
which induces a constant target velocity field
\begin{equation}
u_t = \frac{\mathrm{d}x_t}{\mathrm{d}t} = x_1 - x_0.
\end{equation}
A neural network $v_\theta(x_t, t)$ is trained to approximate this velocity by minimizing the mean squared error
\begin{equation}
\mathcal{L}_{\text{FM}} = \mathbb{E}_{t, x_0, x_1}
\left[ \big\| v_\theta(x_t, t) - u_t \big\|_2^2 \right].
\label{eq.flow_matching}
\end{equation}
The learned velocity field $v_\theta$ implicitly defines a continuous-time dynamical system,
\begin{equation}
\frac{\mathrm{d} x_t}{\mathrm{d} t} = v_\theta(x_t, t),
\end{equation}
with initial condition $x_0 \sim \mathcal{N}(0, I)$.
Samples are obtained by numerically integrating this Ordinary Differential Equation (ODE) from $t=0$ to $t=1$, yielding a mapping from noise to data.

\subsubsection{Shortcut Model}

Standard flow matching learns an instantaneous velocity field, while generation requires numerical integration over multiple
denoising steps. 
To support few-step generation, a shortcut model additionally conditions its prediction on the integration span: $s_\theta(x_t,t,\Delta t)$.
It approximates the average velocity over a finite interval, allowing the state to be updated as
\begin{equation}
x_{t+\Delta t}
\approx
x_t+\Delta t\,s_\theta(x_t,t,\Delta t),
\end{equation}
where $\Delta t>0$ and $t+\Delta t\leq 1$.

Shortcut training exploits a self-consistency relation:
one update of total span $\Delta t$ should agree with two
sequential updates of span $\Delta t/2$.
The intermediate state is predicted by the first half-step:
\begin{equation}
\tilde x_{t+\Delta t/2}
=
x_t+
\frac{\Delta t}{2}
s_\theta(x_t,t,\Delta t/2).
\end{equation}
The target average velocity for the full-span update is
\begin{equation}
\begin{aligned}
s_{\mathrm{target}}(x_t,t,\Delta t)
=\frac{1}{2}\big[
s_\theta(x_t,t,\Delta t/2)
+
s_\theta(\tilde{x}_{t+\Delta t/2},
t+\Delta t/2,\Delta t/2)
\big].
\end{aligned}
\end{equation}
We define the per-example consistency loss as
\begin{equation}
\ell_{\mathrm{SC}}(x_t,t,\Delta t)
=
\left\|
s_\theta(x_t,t,\Delta t)
-
\mathrm{SG}
\left[
s_{\mathrm{target}}(x_t,t,\Delta t)
\right]
\right\|_2^2,
\end{equation}
and the expected consistency objective as
\begin{equation}
\mathcal L_{\mathrm{SC}}
=
\mathbb E_{x_0,x_1,(t,\Delta t)}
\left[
\ell_{\mathrm{SC}}(x_t,t,\Delta t)
\right].
\end{equation}
Here, $\mathrm{SG}[\cdot]$ stops gradients through the self-consistency target.
The intermediate state is generated by the model prediction, rather than by interpolation toward the paired training sample.
At zero integration span, $s_\theta(x_t,t,0)$ represents an instantaneous velocity field and is supervised using the flow-matching objective in Eq.~\ref{eq.flow_matching}.
The standard shortcut model combines flow-matching supervision at zero step size with self-consistency supervision at nonzero step sizes. These two objectives are jointly optimized by sampling different training examples for each objective, yielding
\begin{equation}
\mathcal{L}_{\mathrm{Shortcut}} = \mathcal{L}_{\mathrm{FM}}
+
\lambda\mathcal{L}_{\mathrm{SC}},
\end{equation}
where $\lambda$ controls their relative contribution.

Throughout the preliminaries, $t\in[0,1]$ denotes continuous rectified-flow time, increasing from noise to data, and $\Delta t$ denotes the total integration span. 
For the Wan DiT~\cite{wan} backbone, we define the model timestep before integer quantization as $\tau=N\sigma=N(1-t)$, where
$\sigma$ is the noise coefficient and $N=1000$ is the timestep scaling factor. A denoising transition from $\tau$ to $\tau-\Delta\tau$ has a continuous flow-time span and a noise span satisfying
$\Delta t=\Delta\sigma=\Delta\tau/N$.
The shifted schedule determines the timestep grid used during inference.

\begin{figure*}
\centering{\includegraphics[width=\textwidth]{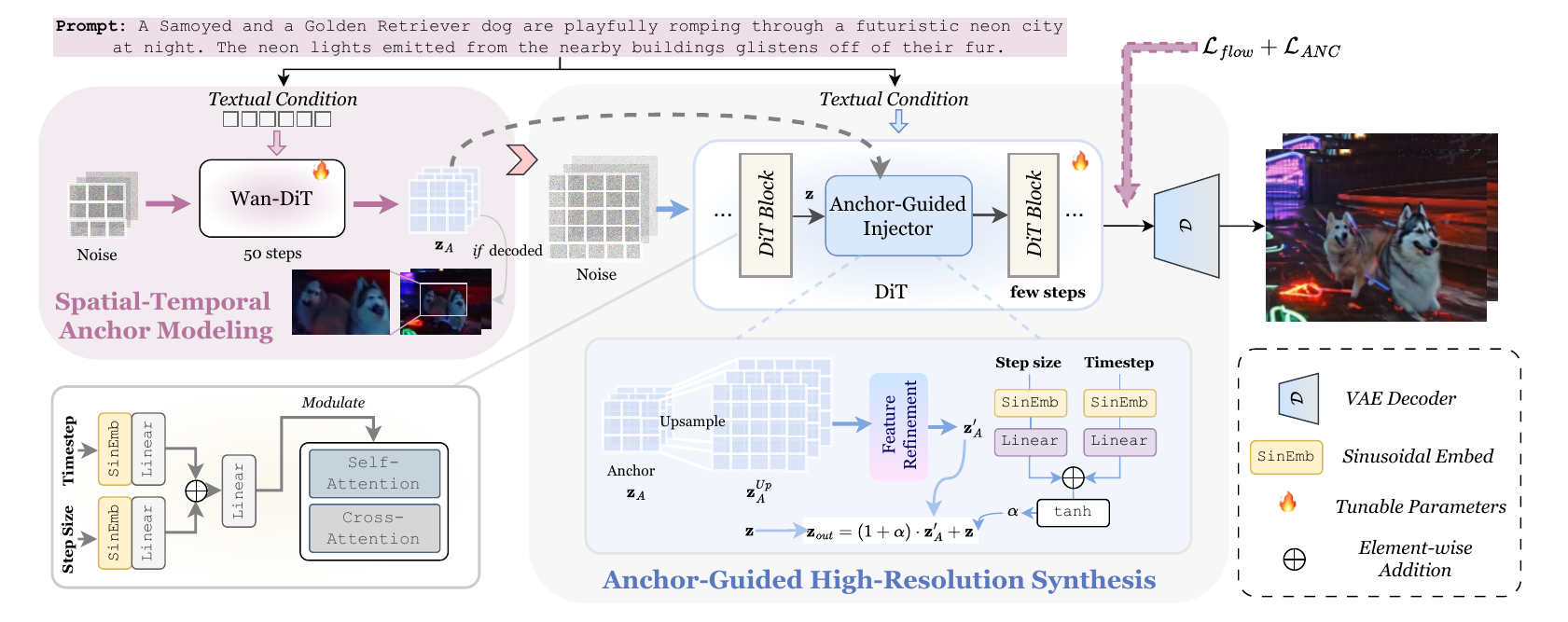}}
\caption{The framework architecture of PixelWizard. We first employ Spatial-Temporal Anchor Modeling to establish a robust global structure using a compact latent representation. Subsequently, Anchor-Guided High-Resolution Synthesis leverages this anchor as a structural constraint to generate high-fidelity textures, ensuring both effective modeling of long-range dependencies and fine-grained detail. Furthermore, we introduce a Noise-Span Aligned Shortcut Training to significantly accelerate inference during the computationally intensive high-resolution stage, enabling efficient detail refinement with minimal sampling steps.}
\label{Fig.Framewok}
\end{figure*}

\subsection{Framework Architecture}
High-resolution video generation faces a semantic density dilemma: as spatial resolution increases, semantic information is distributed over substantially more tokens, making it difficult to jointly model global spatiotemporal structures and fine-grained local textures.
Instead of struggling to learn global spatiotemporal dependencies directly on a semantically sparse high-resolution grid—which leads to optimization difficulties and prohibitive overhead—we adopt a ``divide-and-conquer'' strategy (Fig.~\ref{Fig.Framewok}).
In \emph{Stage I: Spatial-Temporal Anchor Modeling}, we first generate global motion and layout in a compact, high-density latent space. 
By operating in this compact regime, the model can focus on global structures and long-range spatiotemporal dependencies without the burden of high-resolution texture modeling.
In \emph{Stage II: Anchor-Guided High-Resolution Synthesis}, the generated anchor is injected into the DiT backbone as a structural prior via the Anchor-Guided Injector, guiding high-resolution generation while allowing the model to focus its capacity on fine-grained texture synthesis.
By resolving global structure in the anchor modeling stage, the high-resolution stage can focus primarily on detail synthesis under explicit structural guidance, substantially simplifying the learning problem.

\subsection{Spatial-Temporal Anchor Modeling}
To explicitly encode global motion patterns and structural layouts in a high-density latent space, the model is required to operate at a compact, low-resolution regime. We perform Spatial-Temporal
Anchor Modeling at a resolution of $448\times256$.
With a VAE spatial compression factor of $16$ and a spatial patch size of $2\times2$, each frame is represented by a $14\times8$ token grid, containing only $112$ spatial tokens.
However, we observe that models exhibit representation collapse when performing direct inference at substantially lower resolutions. We attribute this failure to a semantic density mismatch in the latent representation. When high-resolution priors are applied to a substantially sparser token grid, each token is forced to encode excessive spatial variation, resulting in object-level structural distortions.

To address this, we perform tuning on the low-resolution data.
Fine-tuning on 65k low-resolution samples from an in-house dataset improves anchor generation over direct low-resolution inference.
This process effectively recalibrates the model to the low-resolution regime, establishing robust spatial-temporal anchors for the subsequent high-resolution synthesis.

\subsection{Anchor-Guided High-Resolution Synthesis}
With the global structural blueprint established by the anchor, the goal of this stage is to synthesize high-fidelity textures coherent with this structure.
To achieve this, we integrate the proposed Anchor-Guided Injector directly into the high-resolution DiT backbone.

\subsubsection{Anchor-Guided Injector}
As shown in Fig.~\ref{Fig.Framewok}, the injector functions as a dynamic interface between the dense structural condition and the high-resolution latent space.
The anchor $\mathbf{z}_A$ is first upsampled and processed by a lightweight Feature Refinement module, consisting of a $1\times1\times1$ convolution followed by GroupNorm and SiLU~\cite{silu}, and a $3\times3\times3$ convolution followed by SiLU.
The refined features are flattened into tokens and normalized using LayerNorm~\cite{layernorm} (LN):
\begin{equation}
\mathbf{z}'_A =
\operatorname{LN}\!\left(
\mathcal{F}_{\mathrm{ref}}\!\left(
\mathcal{U}(\mathbf{z}_A)
\right)\right),
\label{eq:anchor_refinement}
\end{equation}
where $\mathcal{U}$ denotes spatial upsampling and $\mathcal{F}_{\mathrm{ref}}$ encompasses convolutional refinement and flattening into tokens.

We modulate the anchor features using a channel-wise gating mechanism conditioned on the model timestep $\tau$ and timestep span $\Delta\tau$. Their sinusoidal embeddings are processed by two independent MLPs, whose outputs are summed to produce a gating vector:
\begin{equation}
\alpha = \tanh\!\left(
g_{\tau}(\operatorname{SinEmb}(\tau))
+g_{\Delta}(\operatorname{SinEmb}(\Delta\tau))
\right).
\end{equation}
The feature injection is formulated as:
\begin{equation}
\mathbf{z}_{out} = (1 + \alpha) \cdot \mathbf{z}'_A + \mathbf{z},
\end{equation}
where $\mathbf{z}$ denotes the intermediate features of the DiT block. The term $(1 + \alpha)$ acts as a learnable gain.

\subsubsection{Decoupled Training Strategy} 
We adopt a decoupled training paradigm optimized independently after the anchor modeling stage. Specifically, the conditioning anchor is obtained by perturbing the training ground truth via a degradation pipeline, which includes Gaussian blur, spatial resizing, and Gaussian noise. This effectively strips away high-frequency details, compelling the model to leverage its generative priors to hallucinate realistic textures that match the target distribution, preventing trivial reconstruction. 
We jointly optimize the DiT backbone and the anchor-guided injector, allowing the model to recalibrate long-range attention and feature interactions under explicit structural constraints.

\subsection{Noise-Span Aligned Shortcut Training}

While our framework alleviates the optimization difficulty, high-resolution generation remains bottlenecked by inference latency.
Standard flow matching requires many integration steps, while distillation-based methods~\cite{DCM,DMD} incur substantial memory overhead due to teacher--student training.
The shortcut paradigm~\cite{shortcut} provides a distillation-free alternative by learning large-step transitions through temporal self-consistency, enabling few-step inference.
High-resolution generation uses shifted noise schedules~\cite{sd3}, which determine the timestep grid used during inference. With only a few denoising updates, the resulting trajectory includes relatively large noise-span transitions. We define the noise span as the magnitude of the change in noise coefficient across a transition. Uniform sampling over candidate scales allocates equal training frequency to different spans, but does not explicitly prioritize the large-span predictions required for few-step inference.
To place greater training emphasis on large-span transitions, we propose Noise-Span Aligned Shortcut Training, consisting of Exponential Index-Biased Sampling and Adaptive Noise-Span Calibration. The former increases their sampling frequency, while the latter increases their relative contribution to the consistency objective.

\subsubsection{Exponential Index-Biased Sampling}

To increase training exposure to large-span transitions, we introduce biased sampling over candidate scales.
For shortcut consistency training, we select starting model timesteps from $\mathcal A=\{500,600,700,800\}$.
For each selected $\tau\in\mathcal A$, we construct a set of candidate timestep spans covering multiple scales:
\begin{equation}
\mathcal D(\tau)=
\left\{
\Delta\tau_k(\tau)=
\max\!\left(
2,\,
2\left\lfloor\frac{\tau}{2^{k+1}}\right\rfloor
\right)
:\ k=0,\ldots,K-1
\right\}.
\end{equation}
Here, $K=6$, and each span is constrained to be even so that the half-span midpoint lies on an integer model timestep. Smaller candidate indices $k$ correspond to larger spans, yielding
$\Delta\tau_0(\tau)\ge\Delta\tau_1(\tau)\ge
\cdots\ge\Delta\tau_{K-1}(\tau)$.

Uniform sampling assigns equal probability to each candidate scale, without explicitly prioritizing large-span predictions.
To emphasize these predictions, we instead draw the candidate index $k$ from an exponentially decaying distribution:
\begin{equation}
p(k)=
\frac{\exp(-\beta k)}
{\sum_{j=0}^{K-1}\exp(-\beta j)}.
\end{equation}
For $\beta>0$, smaller candidate indices receive higher probabilities, increasing the sampling frequency of larger timestep spans.
The sampled span is $\Delta\tau=\Delta\tau_k(\tau)$.
The corresponding denoising transition moves from $\tau$ to $\tau-\Delta\tau$, with a continuous flow-time span $\Delta t=\Delta\tau/N$.

\subsubsection{Adaptive Noise-Span Calibration}

Exponential Index-Biased Sampling controls how frequently different candidate spans are selected. To complement this sampling bias, Adaptive Noise-Span Calibration adjusts the relative contribution of each example to the consistency objective. Under the timestep convention \(\tau=N\sigma\), the noise span is proportional to the model timestep span. We therefore compute a span-dependent weight directly from \(\Delta\tau\), assigning higher relative weights to larger-span transitions.

For a denoising transition from $\tau$ to $\tau-\Delta\tau$, the noise-span magnitude is
\begin{equation}
\Delta\sigma
=
\frac{\tau}{N}
-
\frac{\tau-\Delta\tau}{N}
=
\frac{\Delta\tau}{N}.
\end{equation}
The corresponding span-dependent weight is
\begin{equation}
w(\Delta\tau)
=
(\Delta\sigma)^p
=
\left(\frac{\Delta\tau}{N}\right)^p,
\end{equation}
where $p=0.5$.
Larger noise spans receive higher relative weights.

Using the per-example consistency loss defined in the preliminaries, the Adaptive Noise-Span Consistency (ANC) objective is
\begin{equation}
\mathcal L_{\mathrm{ANC}}
=
\mathbb E_{\tau,k,x_0,x_1}
\left[
w(\Delta\tau)\,
\ell_{\mathrm{SC}}
\left(x_t,t,\frac{\Delta\tau}{N}\right)
\right],
\end{equation}
where $t=1-\tau/N$ and $\Delta\tau=\Delta\tau_k(\tau)$.
We alternate between $\mathcal L_{\mathrm{FM}}$ and $\mathcal L_{\mathrm{ANC}}$ mini-batches to retain data-grounded instantaneous-velocity supervision while encouraging consistent large-span predictions.

\section{Experiment}

\begin{figure*}[ht]
\centering{\includegraphics[width=\textwidth]{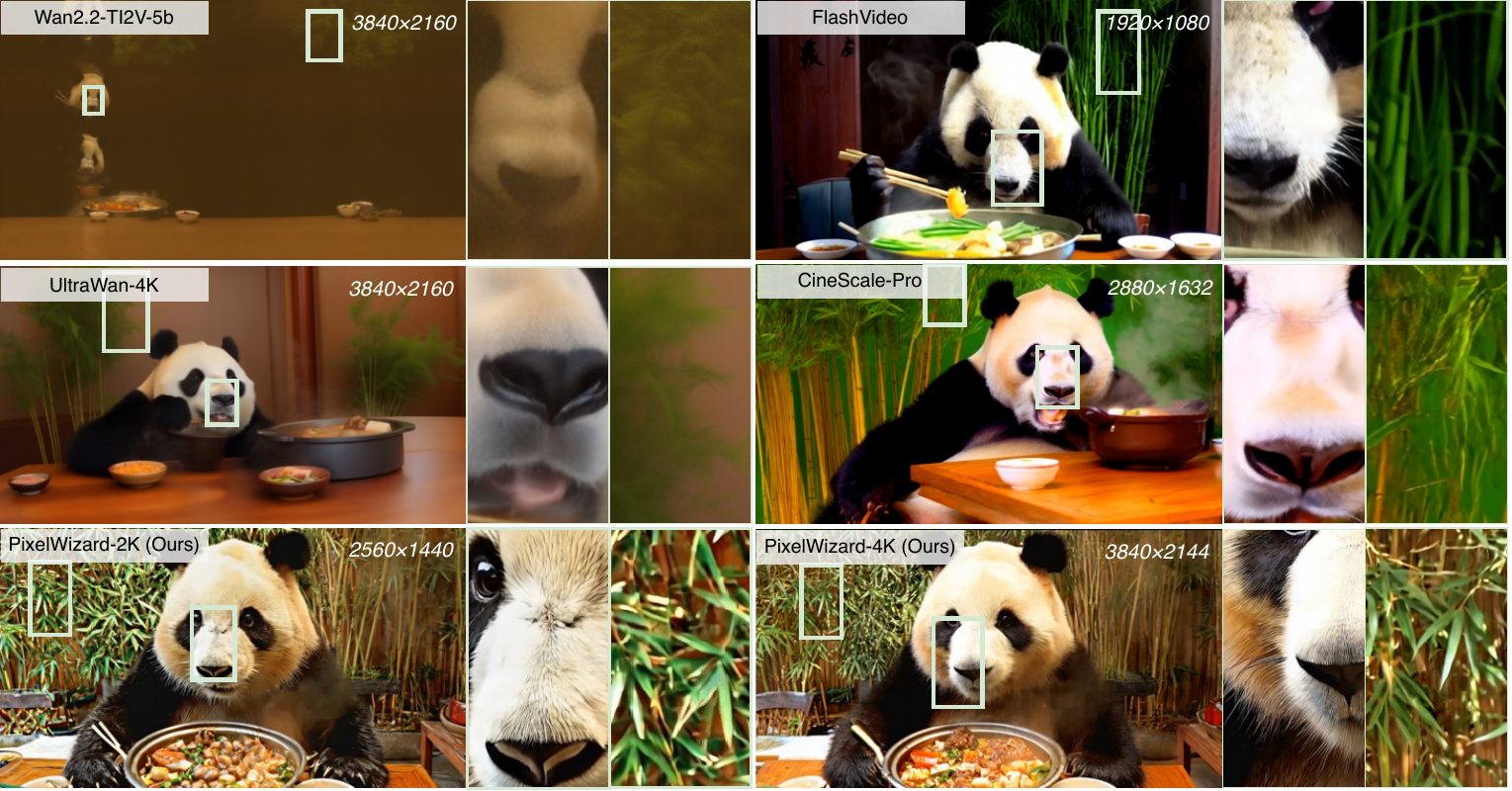}}
\caption{Qualitative comparison with high-resolution video generation methods. We show enlarged views of representative regions (e.g., bamboo and panda) to highlight fine-grained details.}
\label{Fig.Compare_4k}
\vspace{-0.15in}
\end{figure*}

\subsection{Implementation Details}
We adopt Wan2.2-TI2V-5B~\cite{wan} as the base model.
We train \textit{PixelWizard-2K} and \textit{PixelWizard-4K} targeting 2K (2560$\times$1440) and 4K (3840$\times$2144) resolutions, respectively. All experiments are conducted with a video duration of 5 seconds.
Spatial–Temporal Anchor Modeling is performed at a spatial resolution of 448$\times$256.
We use the UltraVideo-4K dataset~\cite{ultravideo}, which contains approximately 42K videos, to train the anchor-guided high-resolution synthesis stage.
Training is conducted using the AdamW~\cite{adamw} optimizer with a learning rate of $1.0\times10^{-5}$, and exponential moving average (EMA) weights are employed to stabilize optimization.
The coefficient $\beta$ is set to 0.7.

\begin{figure*}[!h]
\centering{\includegraphics[width=\textwidth]{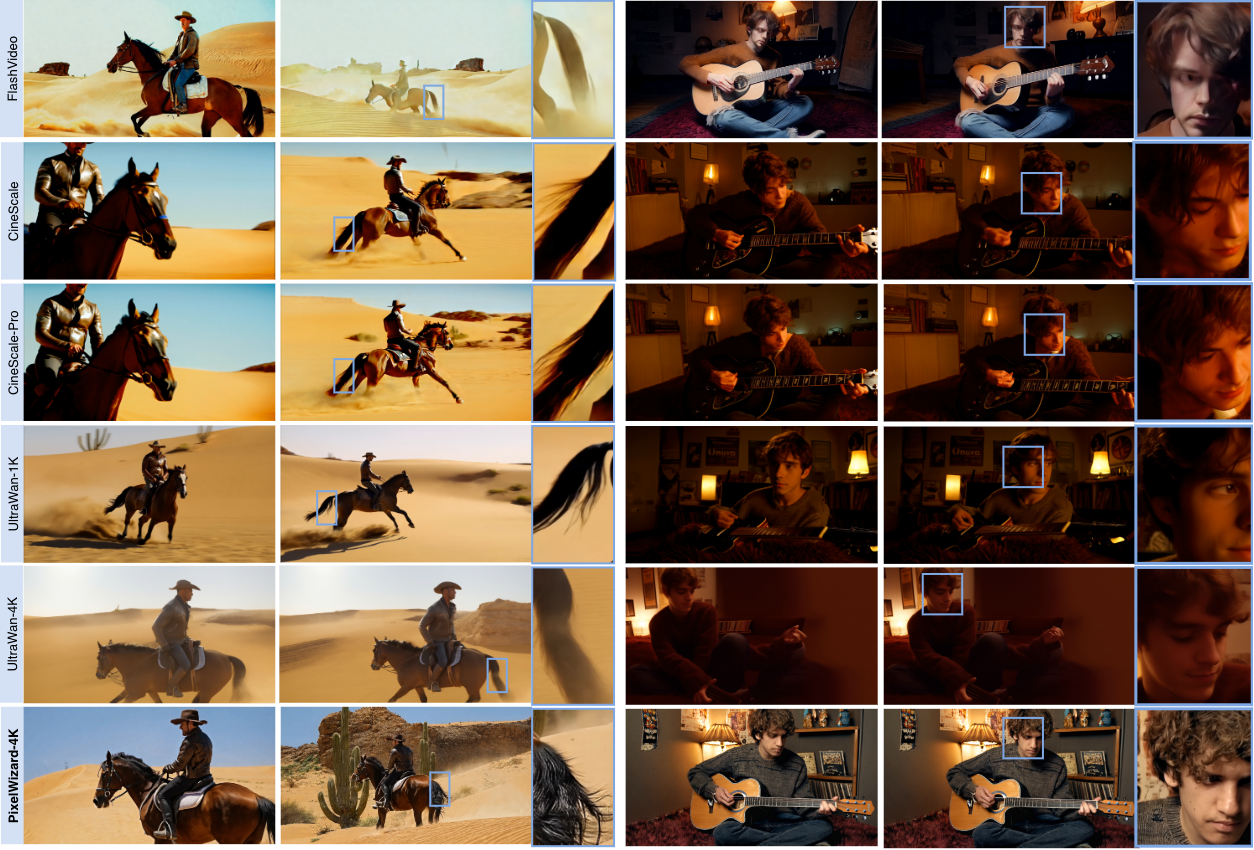}}
\caption{Qualitative comparison with representative high-resolution video generation methods at their supported resolutions. PixelWizard preserves global object structures while producing fine-grained local textures at 4K resolution.}
\label{Fig.apd.compare4}
\end{figure*}

\subsubsection{Compared Methods and Evaluation Protocol}
We compare three representative paradigms:
(i) direct resolution extrapolation of the base model (\textit{e.g.}, Wan2.2-TI2V-5B~\cite{wan});
(ii) training-based high-resolution generation models, including FlashVideo~\cite{FlashVideo} and UltraWan~\cite{ultravideo};
(iii) training-free methods explicitly designed for high-resolution video synthesis, such as CineScale~\cite{CineScale_arxiv2508}.
All methods are evaluated using the same 100 prompts from VBench~\cite{vbench}, following the evaluation protocol in~\cite{FlashVideo}.

\subsubsection{Evaluation Metrics}
We adopt VBench metrics to evaluate complementary aspects of generation quality: Subject Consistency assesses temporal consistency of foreground subjects. Dynamic Degree evaluates dynamics. Aesthetic Quality and Imaging Quality evaluate perceptual quality and visual fidelity, respectively. While these metrics provide a comprehensive evaluation of temporal consistency, motion characteristics, and perceptual quality, they are not specifically designed to assess the detail richness of high-resolution videos.

Building upon prior work on high-resolution video evaluation~\cite{ultragen}, we adopt HD-MSE and HD-LPIPS to evaluate the detail richness and we revise the evaluation protocol to ensure fair comparison across methods operating at different spatial resolutions.
Given a generated video $v \in \mathbb{R}^{T\times C\times H\times W}$, we define a scale-dependent degradation–reconstruction operator $\mathcal{R}_k(\cdot)$ by downsampling the video by a factor of $2^k$ followed by upsampling back to the original resolution.

\begin{itemize}
    \item \textbf{mHD-MSE.}
we use mHD-MSE to evaluate the preservation of fine-grained details in high-resolution videos. Specifically, a generated video is downsampled by factors of $2^k$, producing a set of downsampled videos, which are then upsampled back to the original resolution. The mean squared error between the reconstructed and original videos is accumulated across scales:
\begin{equation}
\text{mHD-MSE}(v)
=
\frac{1}{|\mathcal{K}|}
\sum_{k\in\mathcal{K}}
\frac{1}{TCHW}
\left\|
v - \mathcal{R}_{k}(v)
\right\|_2^2,
\end{equation}
We follow the standard setting in~\cite{ultragen} and use $k=\{3,4,5\}$, corresponding to downsampling factors of 8, 16, and 32.
\item \textbf{mHD-LPIPS.}
Analogous to HD-MSE, HD-LPIPS replaces the pixel-wise MSE with the perceptual LPIPS metric to better capture semantic and perceptual differences in high-resolution content. The metric is computed as:
\begin{equation}
\text{mHD-LPIPS}(v)
=
\frac{1}{|\mathcal{K}|}
\sum_{k \in \mathcal{K}}
\frac{1}{T}
\sum_{t=1}^{T}
\text{LPIPS}\!\left(
v_t,\,
\mathcal{R}_k(v)_t
\right)
,
\end{equation}
with $k=\{3,4,5\}$.
\end{itemize}

mHD-MSE and mHD-LPIPS quantify the amount of high-frequency information that is lost under scale-space degradation.
Therefore, higher values indicate richer fine-grained details, as videos with more complex textures suffer larger discrepancies after downsampling and reconstruction.

\begin{table*}[ht]
\centering
\setlength{\tabcolsep}{1mm}{
\caption{Quantitative comparison of high-resolution video generation across different resolutions, evaluating high-frequency detail (mHD-MSE and mHD-LPIPS), perceptual quality (DOVER~\cite{DOVER}), and multidimensional video quality (VBench~\cite{vbench}). \textbf{Bold} and \underline{underlined} values indicate the best and second-best performances within each resolution category, respectively.}
\label{Tab.compare_huazhi}
\scalebox{0.7}{
\begin{tabular}{lc|cc|cc|ccccc} 
\Xhline{1.pt}
\textbf{Models} &  
\textbf{Resolution} & 
\textbf{mHD-MSE$\uparrow$} & 
\textbf{mHD-LPIPS$\uparrow$} & 
\textbf{Tech.$\uparrow$} & 
\textbf{Aesth.$\uparrow$} &
\makecell{\textbf{Subject}\\ \textbf{Consistency}} &
\makecell{\textbf{Dynamic}\\ \textbf{Degree}} & 
\makecell{\textbf{Aesthetic}\\ \textbf{Quality}} & 
\makecell{\textbf{Imaging}\\ \textbf{Quality}}  & \makecell{\underline{\textbf{Avg.}}} \\
\midrule
FlashVideo~\cite{FlashVideo} & $1920\times1080$  & 0.0098 & 0.4756 & 12.47 & 98.11 & 94.24 & \underline{51.00} & 61.01 & \underline{71.05} & 69.33\\
UltraWan-1k~\cite{ultravideo} & $1920\times1088$ & 0.0069 & 0.4517 & 13.77 & 98.73 & \textbf{96.51} & 25.00 & \textbf{65.09} & 69.83 & 64.11\\
CineScale~\cite{CineScale_arxiv2508} & $1920\times1088$   & 0.0067 & 0.3631 & 12.83 & 98.21 & 93.19 & \textbf{54.00} & 63.26 & 68.67 & \underline{69.78}\\
Wan2.2-TI2V-5B~\cite{wan} & $2560\times1440$ & 0.0052 & 0.2889 & 12.98 & 98.01 & 93.25 & 26.00 & 63.59 & 69.34 & 63.05\\
\midrule
\rowcolor{cyan!10} \textbf{PixelWizard-2K}& $2560\times1440$  & \textbf{0.0147} & \textbf{0.5206} &  \textbf{14.31} & \textbf{99.58} & \underline{96.16} & 49.00 & \underline{64.39} & \textbf{72.64} & \textbf{70.55}\\
\midrule
\Xhline{1.pt}
\textbf{Models} &  
\textbf{Resolution} & 
\textbf{mHD-MSE$\uparrow$} & 
\textbf{mHD-LPIPS$\uparrow$} & 
\textbf{Tech.$\uparrow$} & 
\textbf{Aesth.$\uparrow$} &
\makecell{\textbf{Subject}\\ \textbf{Consistency}} &
\makecell{\textbf{Dynamic}\\ \textbf{Degree}} & 
\makecell{\textbf{Aesthetic}\\ \textbf{Quality}} & 
\makecell{\textbf{Imaging}\\ \textbf{Quality}}  & \makecell{\underline{\textbf{Avg.}}} \\
\midrule
CineScale-Pro~\cite{CineScale_arxiv2508} & $2880\times1632$ & 0.0050 & 0.3483 & 12.76 & 98.15 & 92.96 & \textbf{56.00} & \underline{64.05} & \underline{69.58} & \underline{70.65}\\
UltraWan-4k~\cite{ultravideo} & $3840\times2160$ & 0.0012 & 0.3319 & 10.59 & 98.88 & \textbf{98.64} & 10.00 & 60.88 & 61.30 & 57.71\\
Wan2.2-TI2V-5B~\cite{wan} & $3840\times2144$ & 0.0010 & 0.3276 & 9.89 & 97.99 & 93.16 & 32.00 & 46.11 & 46.96 & 54.56\\
\midrule
\rowcolor{cyan!10} \textbf{PixelWizard-4K} & $3840\times2144$ & \textbf{0.0102} & \textbf{0.5120} & \textbf{14.14} & \textbf{99.66} & \underline{96.25} & \underline{49.00} & \textbf{64.43} & \textbf{74.25} & \textbf{70.98}\\
\Xhline{1pt}
\end{tabular}}}
\end{table*}

\subsection{Comparison with State-of-the-Art Methods}

\subsubsection{Qualitative Comparison with High-Resolution Generation Methods}

We present qualitative comparisons with high-resolution video generation methods in Figs.~\ref{Fig.Compare_4k} and~\ref{Fig.apd.compare4}.
To examine fine-grained visual quality, we show enlarged views of representative regions.
As shown in Fig.~\ref{Fig.Compare_4k}, directly applying Wan2.2-TI2V-5B at 4K produces poorly resolved foreground content. 
Although FlashVideo and UltraWan-4K generate recognizable scene content, their enlarged regions reveal limitations in local structure and texture.
The bamboo foliage exhibits less distinct leaf boundaries, while the panda's face contains softened textures and less clearly resolved fur detail.
CineScale-Pro presents a clearer overall scene, yet its enlarged regions still lack well-separated fine textures.
By comparison, PixelWizard delineates individual leaves more clearly and produces more visible fur texture around the panda's face, together with better-defined facial contours.
Fig.~\ref{Fig.apd.compare4} shows that these detail differences also appear in full scenes. FlashVideo, CineScale, and CineScale-Pro depict the horse-riding and guitar-playing subjects, but their enlarged horse-tail or facial regions contain comparatively limited texture. UltraWan maintains recognizable scene layouts, while its 4K crops appear noticeably soft. PixelWizard-4K retains the rider–horse and guitarist–guitar structures and resolves finer texture in the selected crops.
These improvements are apparent in both the background foliage and the foreground subject, illustrating richer fine-scale detail alongside clearer local structures in the displayed examples.

\subsubsection{Quantitative Comparison with High-Resolution Generation Methods}

Table~\ref{Tab.compare_huazhi} evaluates high-resolution detail richness using mHD-MSE and mHD-LPIPS, perceptual quality using the technical and aesthetic scores of DOVER~\cite{DOVER}, and multidimensional video quality using VBench~\cite{vbench}.
PixelWizard consistently achieves the highest mHD-MSE and mHD-LPIPS scores across both 2K and 4K resolution categories, demonstrating its effectiveness in generating rich fine-grained visual details.
At 2K resolution, PixelWizard achieves an mHD-MSE of 0.0147 and an mHD-LPIPS of 0.5206, outperforming all competing methods.
More importantly, this advantage becomes particularly pronounced at 4K resolution, where existing approaches exhibit substantial degradation in high-frequency detail metrics.
For example, the mHD-LPIPS score of UltraWan decreases by 26.5\% when moving from 2K to 4K, whereas PixelWizard exhibits only a marginal reduction of 1.7\% (from 0.5206 to 0.5120).
These results highlight the ability of PixelWizard to maintain rich high-frequency details even under challenging ultra-high-resolution generation settings.

Beyond detail richness, PixelWizard also consistently achieves the highest DOVER technical and aesthetic scores in both resolution categories.
Specifically, PixelWizard-2K obtains technical and aesthetic scores of 14.31 and 99.58, respectively, while PixelWizard-4K achieves 14.14 and 99.66.
Notably, at 4K resolution, the technical quality scores of UltraWan and Wan2.2 decrease to 10.59 and 9.89, respectively, whereas PixelWizard maintains a substantially higher score of 14.14.
These results suggest that the increased high-frequency information does not come at the expense of perceptual quality.
Instead, PixelWizard effectively balances fine-grained detail enhancement with overall visual fidelity, achieving superior technical quality and aesthetic appeal.

The VBench evaluation further assesses the generated videos from multiple perspectives, including subject consistency, dynamic degree, aesthetic quality, and imaging quality.
PixelWizard achieves the highest average scores in both resolution categories, reaching 70.55 at 2K and 70.98 at 4K.
In particular, PixelWizard demonstrates a clear advantage in imaging quality, achieving scores of 72.64 and 74.25 at 2K and 4K resolutions, respectively.
At 4K resolution, PixelWizard surpasses UltraWan-4K and Wan2.2-TI2V-5B by 12.95 and 27.29 points in imaging quality, respectively.
Meanwhile, PixelWizard maintains strong subject consistency (96.16 and 96.25) and competitive dynamic degree (49.00 at both resolutions), indicating that its improvements in visual fidelity are accompanied by stable subject appearance and substantial motion.
Furthermore, PixelWizard-4K achieves the highest aesthetic quality score of 64.43, reinforcing its advantage in generating visually appealing high-resolution videos.
Overall, these results demonstrate that PixelWizard enhances fine-grained visual details while maintaining strong perceptual quality and balanced video generation performance across multiple evaluation dimensions.

\begin{table}
\centering
\caption{Quantitative comparison of video generation quality and efficiency across resolutions. Inference latency is measured for generating 5-second videos on a single NVIDIA H200 GPU.}
\label{compare_single}

\setlength{\tabcolsep}{4mm}
\scalebox{0.9}{
\begin{tabular}{lc|ccc}
\Xhline{1pt}
\textbf{Models} &
\makecell{\textbf{Resolution}\\($H\times W$)} &
\makecell{\textbf{Latency}\\\textit{(sec)}} &
\makecell{\textbf{Latency}\\\textbf{per Pixel}$\downarrow$} &
\makecell{\textbf{Latency}\\\textbf{per Frame}$\downarrow$} \\
\midrule

FlashVideo~\cite{FlashVideo}
& $1920\times1072$
& 124
& 1.23e-6
& 2.53 \\

UltraWan-1k~\cite{ultravideo}
& $1920\times1088$
& 1,623
& 9.66e-6
& 20.04 \\

CineScale~\cite{CineScale_arxiv2508}
& $1920\times1088$
& 3,138
& 1.85e-5
& 38.74 \\

Wan2.2-TI2V-5B~\cite{wan}
& $2560\times1440$
& 1,635
& 3.66e-6
& 13.51 \\
\midrule
\rowcolor{cyan!10} \textbf{PixelWizard-2K}
& $2560\times1440$
& 165
& \textbf{3.27e-7}
& \textbf{1.21} \\

\midrule
\Xhline{1pt}
\textbf{Models} &
\makecell{\textbf{Resolution}\\($H\times W$)} &
\makecell{\textbf{Latency}\\\textit{(sec)}} &
\makecell{\textbf{Latency}\\\textbf{per Pixel}$\downarrow$} &
\makecell{\textbf{Latency}\\\textbf{per Frame}$\downarrow$} \\
\midrule

CineScale-Pro~\cite{CineScale_arxiv2508}
& $2880\times1632$
& 15,235
& 4.00e-5
& 188.09 \\

UltraWan-4k~\cite{ultravideo}
& $3840\times2160$
& 24,125
& 3.59e-5
& 298.17 \\

Wan2.2-TI2V-5B~\cite{wan}
& $3840\times2144$
& 7,150
& 7.18e-6
& 59.09 \\
\midrule

\rowcolor{cyan!10} \textbf{PixelWizard-4K}
& $3840\times2144$
& 590
& \textbf{5.92e-7}
& \textbf{4.88} \\

\Xhline{1pt}
\end{tabular}
}
\end{table}
\begin{table}
\centering
\setlength{\tabcolsep}{1mm}{
\caption{Quantitative comparison with representative video super-resolution (VSR) methods.} 
\label{compare_vsr}
\scalebox{1}{
\begin{tabular}{lc|cccc|cc} 
\Xhline{1.pt}
\textbf{Models} &  \makecell{\textbf{Resolution}\\ (H$\times$W)}  & \textbf{MUSIQ$\uparrow$} & \textbf{NIQE$\downarrow$} & \textbf{mHD-MSE$\uparrow$}  & \textbf{mHD-LPIPS$\uparrow$} & \textbf{Tech.}$\uparrow$ &
\textbf{Aesth.}$\uparrow$\\

\midrule
STAR~\cite{STAR}     & $2240\times1280$ & 37.80 & 5.55 & 0.0040 & 0.3744 & 11.47 & 98.75\\
DOVE~\cite{DOVE_Nips25}     & $2176\times1280$ & 50.54 & 4.92 & 0.0050 & 0.3943 & 12.31 & 99.21\\
SeedVR2~\cite{Seedvr2_arxiv2025}  & $2528\times1440$ & 40.45 & 4.95 & 0.0063 & 0.3922 & 10.61 & 99.07 \\
FlashVSR~\cite{Flashvsr_Arxiv25} & $2176\times1280$ & 42.84 & 5.53 & 0.0076 & 0.4946 & 13.52 & 99.36\\
\midrule
\rowcolor{cyan!10} \textbf{PixelWizard-2k}& $2560\times1440$  & \textbf{57.67} & \textbf{3.94} & \textbf{0.0147} & \textbf{0.5206} & \textbf{14.31} & \textbf{99.58}\\
\midrule
\Xhline{1.pt}
\textbf{Models} &  \makecell{\textbf{Resolution}\\ (H$\times$W)}  & \textbf{MUSIQ$\uparrow$} & \textbf{NIQE$\downarrow$} & \textbf{mHD-MSE$\uparrow$}  & \textbf{mHD-LPIPS$\uparrow$} & \textbf{Tech.}$\uparrow$ &
\textbf{Aesth.}$\uparrow$\\
\midrule
STAR~\cite{STAR}     & $3584\times2048$ & 20.98 & 7.42 & 0.0015 & 0.2115 & 6.59 & 97.31 \\
DOVE~\cite{DOVE_Nips25}     & $3584\times2048$ & 37.81 & 6.18 & 0.0030 & 0.3094 & 10.37 & 99.15\\
FlashVSR~\cite{Flashvsr_Arxiv25} & $3584\times2048$ & 43.93 & 4.32 & 0.0053 & 0.4437 & 12.80 & 99.38\\
\midrule
\rowcolor{cyan!10} \textbf{PixelWizard-4k} & $3840\times2144$ & \textbf{48.17} & \textbf{4.19} & \textbf{0.0102} & \textbf{0.5120} & \textbf{14.14} & \textbf{99.66}  \\
\Xhline{1pt}
\end{tabular}}}
\end{table}

\begin{figure*}
\centering{\includegraphics[width=\textwidth]{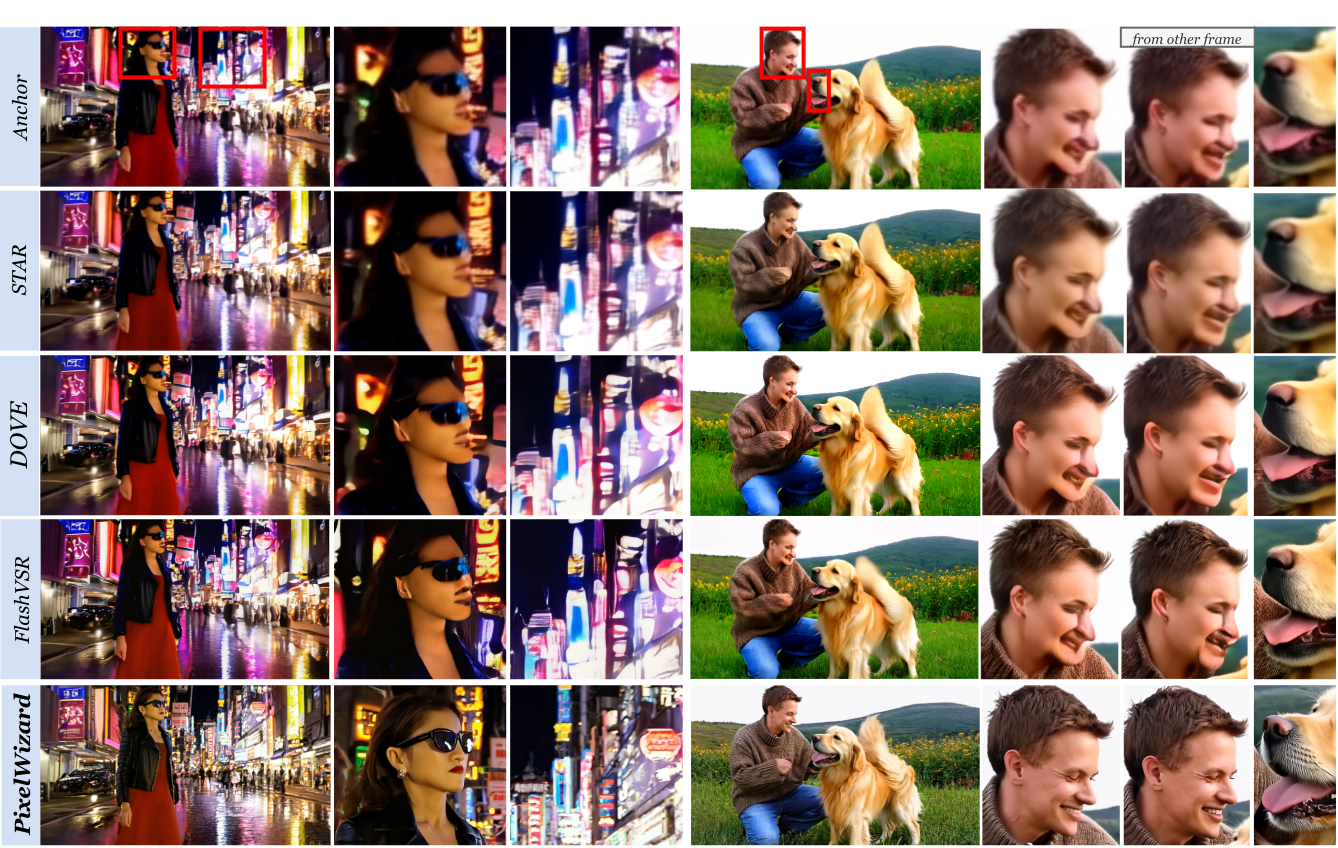}}
\caption{Comparison with video SR methods at 4K resolution. We visualize zoomed-in regions to highlight details.}
\label{Fig.Compare_vsr}
\end{figure*}

\subsubsection{Inference Efficiency Analysis}
Table~\ref{compare_single} summarizes the runtime efficiency of different video generation models measured for the diffusion inference process.
Efficiency is further evaluated using latency normalized by spatial resolution and frame count.
Several methods incur extremely large latency when scaled to 4K resolution, with generating a single short video requiring hours of computation. The excessive inference cost severely limits practical deployment in real-world content creation pipelines.
In contrast, PixelWizard maintains strong efficiency at both 2K and 4K resolutions, enabling high-resolution video synthesis within a practical time budget.

\subsubsection{Comparison with Video Super-Resolution Methods}
We further compare PixelWizard with generative video super-resolution (VSR) pipelines, where a low-resolution video is first generated and subsequently upsampled to the target resolution using STAR~\cite{STAR}, DOVE~\cite{DOVE_Nips25}, SeedVR2~\cite{Seedvr2_arxiv2025}, and FlashVSR~\cite{Flashvsr_Arxiv25}.
In Table~\ref{compare_vsr}, we additionally report the no-reference image quality metrics MUSIQ~\cite{ke2021musiq} and NIQE~\cite{NIQE}, with qualitative comparisons provided in Fig.~\ref{Fig.Compare_vsr}.
Despite their effectiveness in enhancing visual details, VSR-based approaches are fundamentally constrained by the content and structures established in the low-resolution inputs. As post-processing methods, they primarily focus on recovering high-frequency details rather than correcting underlying structural or semantic deficiencies. Consequently, inaccuracies or missing information in the low-resolution generation may propagate to the enhanced results, leading to structural distortions and temporal inconsistencies. In contrast, PixelWizard treats the low-resolution representation as a structural anchor rather than a fixed reconstruction target, enabling direct content synthesis in the high-resolution latent space. This design preserves global spatiotemporal coherence while allowing the model to generate fine-grained details and high-resolution content beyond the information explicitly available in the low-resolution anchor.

\begin{table}[!h]
\centering
\setlength{\tabcolsep}{1.2mm}{
\caption{Parameter count (Params.) and peak memory comparison on a single H200 GPU. OOM denotes out-of-memory errors caused by insufficient GPU memory.} 
\label{Tab.appendix_param_mem}
\scalebox{1}{
\begin{tabular}{l|cccc} 
\Xhline{1.pt}
\textbf{Method}  & \textbf{Params.} & \textbf{Peak Memory} \\
\midrule
FlashVideo (adapted for 4K inference)   & 7B & OOM  \\
UltraWan-4K  & 1.3B & OOM \\
\midrule
\textbf{PixelWizard-4K} & 10B & 101.8GB\\
\Xhline{1pt}
\end{tabular}}}
\end{table}

\subsubsection{Parameter and Memory Comparison}
We compare the methods with FlashVideo~\cite{FlashVideo} (adapted for 4K inference) and UltraWan-4K~\cite{ultravideo}. 
PixelWizard reports the total parameters across its two sequential stages.
As shown in Table~\ref{Tab.appendix_param_mem}, although our two-stage design introduces extra complexity, it is a deliberate choice for stable 4K generation to ensure stable performance. Despite this, PixelWizard remains relatively efficient and supports 4K generation on a single GPU, making it practical for research and deployment.

\begin{figure*}
\centering{\includegraphics[width=\linewidth]{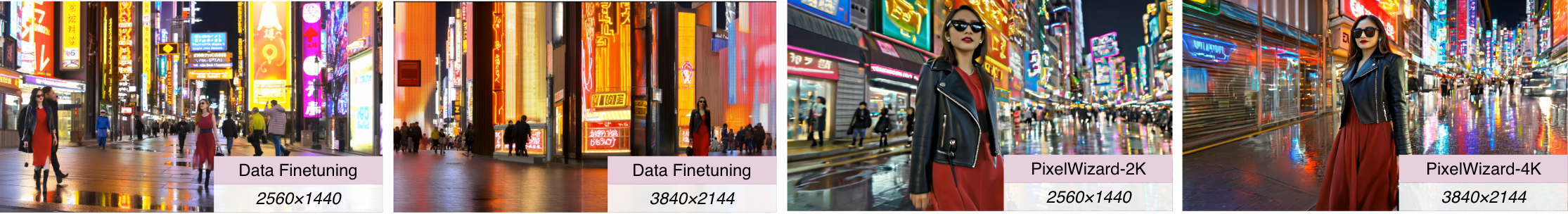}}
\caption{Comparison with direct high-resolution fine-tuning under the same number of training iterations at 2K and 4K resolutions. Direct fine-tuning exhibits structural degradation, including repeated subjects and unstable spatial layouts, whereas PixelWizard better preserves global structure by decoupling structural modeling from high-resolution detail synthesis.}
\label{Fig.Compare_finetune}
\end{figure*}
\begin{figure}
\centering{\includegraphics[width=0.6\linewidth]{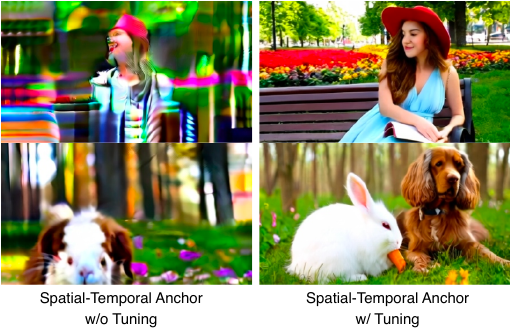}}
\caption{Visual comparison between tuned and untuned anchor models.}
\label{Fig.Ablation_anchor}
\end{figure}

\begin{figure*}[t]
\centering{\includegraphics[width=0.99\textwidth]{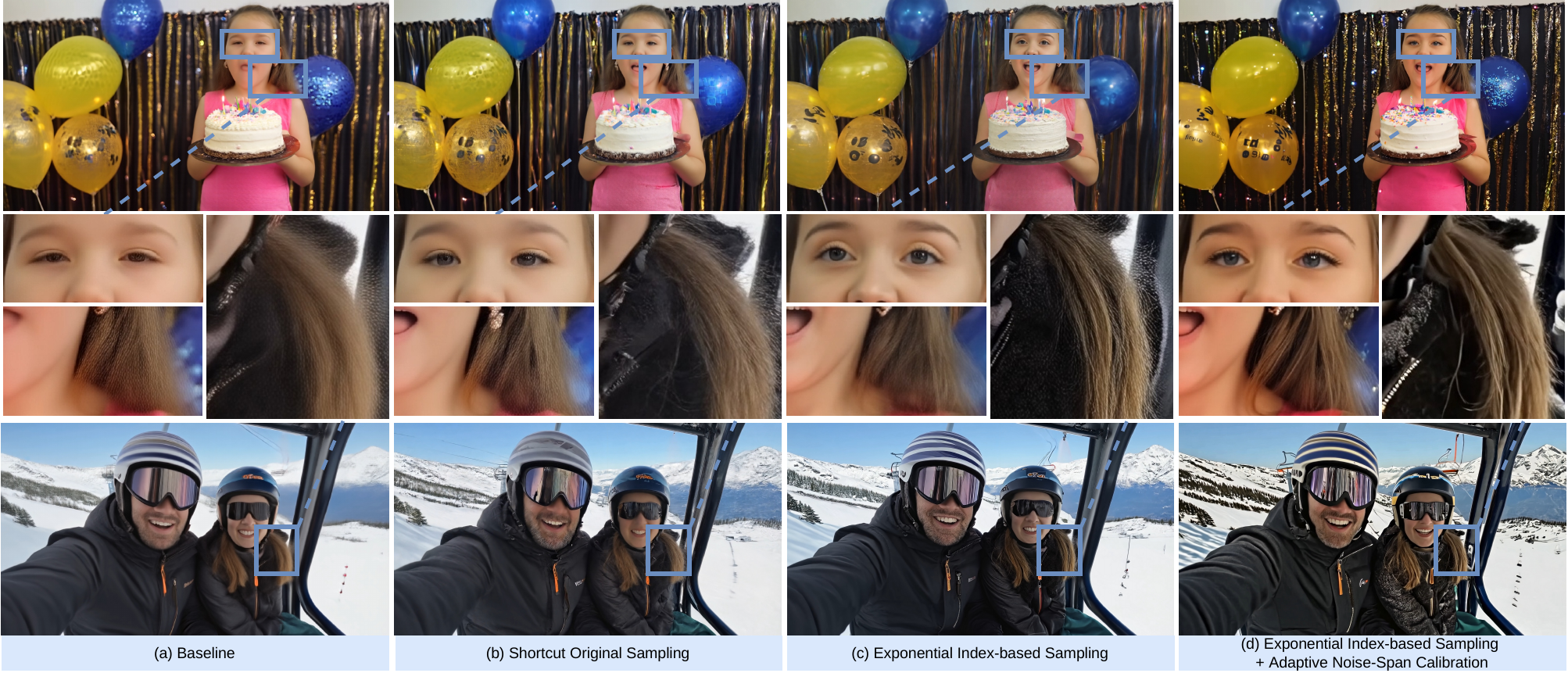}}
\caption{Visual ablation of the proposed training Noise-Span Aligned Shortcut Training strategy under 4-step inference. Variants in (a)–(c) suffer from blurred appearances and inferior hair detail compared to the full model. (Please zoom in for better visualization)}
\label{Fig.Ablation_shortcut}
\end{figure*}

\subsection{Ablation Study}

\subsubsection{Comparison with Direct High-Resolution Fine-tuning}
We further compare our approach with direct fine-tuning at 2K and 4K resolutions using the same number of training iterations, which is shown in Fig.~\ref{Fig.Compare_finetune}.
Direct fine-tuning suffers from repeated subjects, spatial misalignment, and unstable layouts, suggesting that learning global structures directly in the spatially expanded HR latent space is highly data-demanding.
In contrast, our approach first establishes layout and motion in a compact latent space and then uses them as structural anchors for high-resolution synthesis.
This decomposition concentrates global structure learning in a semantically denser representation, enabling more coherent structures and superior visual quality with limited high-resolution training data.

\subsubsection{Effect of Training Spatial-Temporal Anchor}
We designate $448{\times}256$ as the resolution for Spatial-Temporal Anchor Modeling.
With a VAE spatial compression ratio of $16\times$ and a patch size of $2\times$, each frame is represented by a $14{\times}8$ token grid, containing only 112 spatial tokens.
This compact representation substantially reduces the spatial modeling burden while retaining sufficient granularity to capture global layout and motion; further reducing the resolution may compromise spatial topology.
We further investigate whether explicit low-resolution tuning is necessary for learning effective Spatial-Temporal Anchors, as shown in Fig.~\ref{Fig.Ablation_anchor}.
As shown in Table~\ref{Tab.ablation_HR_stage}, anchor tuning improves the technical score from 3.30 to 8.01 and the aesthetic score from 79.11 to 97.91.
Directly applying the pretrained model at $448{\times}256$ results in severe structural degradation, including distorted layouts and unstable motion, indicating a mismatch between its learned high-resolution prior and the compact token space.
Fine-tuning at the target resolution adapts the model to this compact representation, enabling it to reliably capture the global structure and motion required for subsequent high-resolution synthesis.

\subsubsection{Anchor-Guided High-resolution Synthesis}
As shown in Table~\ref{Tab.ablation_HR_stage}, directly injecting the anchor through element-wise addition (\textit{w/o AGI}) yields suboptimal performance, as it overlooks the representation gap between compact structural anchors and high-resolution features.
The Anchor-Guided Injector effectively bridges this gap by aligning and integrating anchor information into the synthesis process.
Further introducing adaptive gating consistently improves performance by dynamically controlling the anchor contribution across denoising stages, enabling more flexible integration of structural guidance during high-resolution synthesis.

\begin{figure}
\centering{\includegraphics[width=0.5\linewidth]{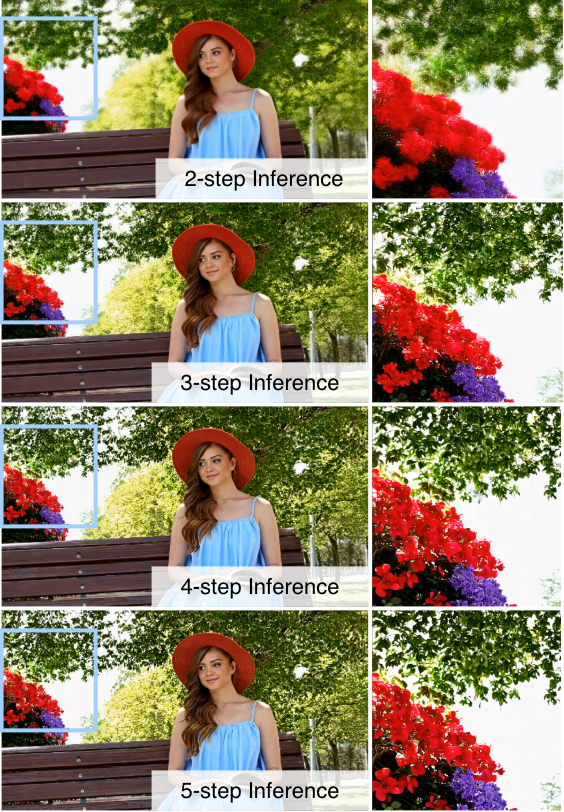}}
\caption{Comparison with different inference steps.}
\label{Fig.Compare_steps}
\end{figure}
\begin{table}[htbp]
    \centering
    \caption{Ablation on Anchor-Guided High-Resolution Synthesis (AGHS) at 2560*1440 resolution. ``Anchor w/o Tuning'' denotes direct low-resolution inference without anchor-stage adaptation; ``w/o AGI'' replaces the proposed Anchor-Guided Injector with direct feature addition; and $\alpha$=0 fixes the adaptive gating term to zero while retaining anchor injection.}
    \label{Tab.ablation_HR_stage}
    \scalebox{1}{
        \setlength{\tabcolsep}{15pt}
        \begin{tabular}{l|cc}
        \Xhline{1pt}
        \textbf{Component} & \textbf{Tech. Score$\uparrow$} & \textbf{Aesth. Score$\uparrow$} \\
        \midrule
        Anchor & 8.01 & 97.91 \\
        Anchor w/o Tuning & 3.30 & 79.11 \\
        \midrule
        +AGHS w/o AGI & 12.51 & 98.99 \\
        +AGHS ($\alpha$=0) & 13.47 & 99.36 \\
        +AGHS & 14.31 & 99.58 \\
        \Xhline{1pt}
        \end{tabular}
    }
\end{table}


\begin{figure}[ht]
\centering{\includegraphics[width=0.6\linewidth]{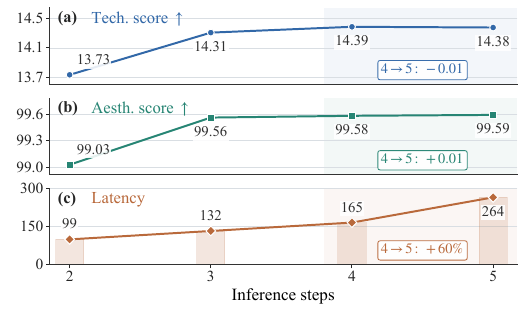}}
\caption{Ablation study of the number of inference steps in the Anchor-Guided High-Resolution Synthesis stage at $2560 \times 1440$ resolution. (a) Technical score, (b) aesthetic score, and (c) latency. }
\label{Fig.ablation_inference_steps}
\end{figure}

\subsubsection{Noise-Span Aligned Shortcut Training}
Fig.~\ref{Fig.Ablation_shortcut} compares different shortcut training strategies.
The baseline (a) exhibits severe over-smoothing, while standard Shortcut training (b) improves local details but still produces blurred fine textures.
Compared with the shortcut baseline, Exponential Index-Biased Sampling improves fine-grained detail, supporting the benefit of increasing training exposure to larger-span transitions.
Exponential Index-Biased Sampling (c) substantially restores high-frequency details by emphasizing these transitions.
Further incorporating Adaptive Noise-Span Calibration yields the best visual fidelity by adjusting the optimization weight according to the physical noise change of each transition.
These results support the effectiveness of combining biased span sampling with span-dependent consistency weighting for few-step generation.

\subsubsection{Inference Steps}
We further analyze the effect of inference steps in Fig.~\ref{Fig.Compare_steps} and Fig.~\ref{Fig.ablation_inference_steps}.
With only 2 steps, the generated results exhibit noticeable blurring in high-frequency regions.
Increasing the number of steps progressively improves fine-grained details, with the quality largely saturating at 4 steps.
A further increase to 5 steps brings only marginal improvements in both perceptual quality and quantitative metrics.
We therefore adopt 4-step inference as the default setting, providing a favorable trade-off between generation quality and efficiency.
\section{Conclusion}
In this work, we present PixelWizard, a framework that explicitly decouples spatial-temporal structure modeling from high-resolution synthesis. 
By resolving global dynamics via Spatial-Temporal Anchor Modeling in a compact latent space and integrating them through an Anchor-Guided Injector, our approach ensures both structural coherence and high-fidelity details. 
Furthermore, the proposed Noise-Span Aligned Shortcut Training enables efficient few-step inference, successfully bypassing the memory bottlenecks of conventional distillation. 
Extensive experiments demonstrate that PixelWizard significantly improves generation quality and efficiency at 2K/4K resolutions, offering a scalable solution for high-resolution video generation.

\bibliography{ref}

\clearpage
\renewcommand{\thefigure}{\thesection.\arabic{figure}}
\renewcommand{\thetable}{\thesection.\arabic{table}}
\renewcommand{\theequation}{\thesection.\arabic{equation}}
\setcounter{figure}{0}
\setcounter{table}{0}
\setcounter{equation}{0}

This is supplementary material for \textbf{\textit{PixelWizard: Towards Efficient High-Fidelity Video Generation at Ultra-Large Spatial Resolutions.}}

\section{Additional Experiments}
\label{Sec.C}

\subsection{More Visual Comparison with Different Models}
We provide additional qualitative comparisons across different video generation models.
Our method yields more consistent structures and finer visual details.

\begin{figure*}
\centering{\includegraphics[width=0.9\textwidth]{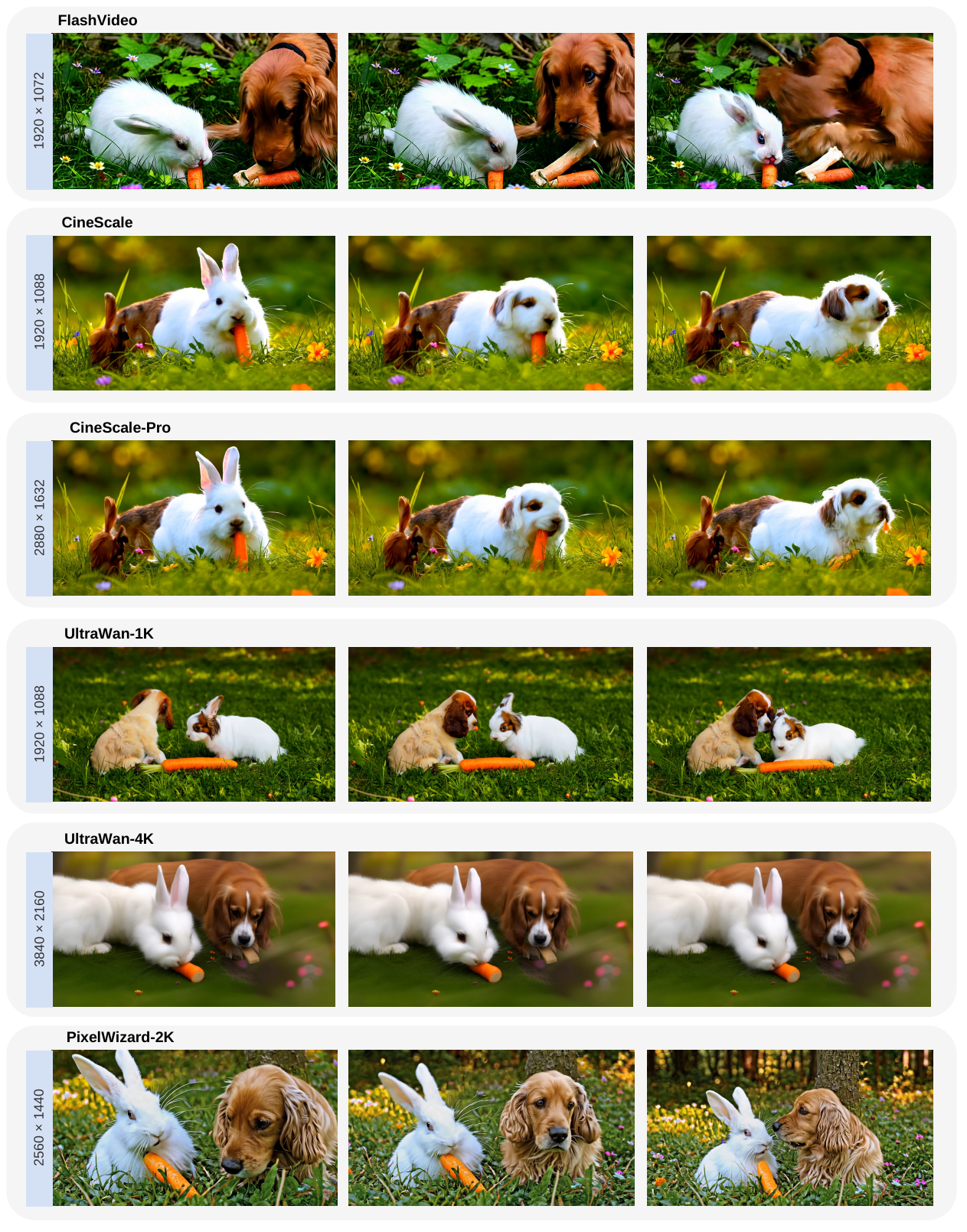}}
\caption{Visual comparison with high-resolution video generation methods. }
\label{Fig.apd.compare1}
\end{figure*}
\begin{figure}
\centering{\includegraphics[width=0.99\linewidth]{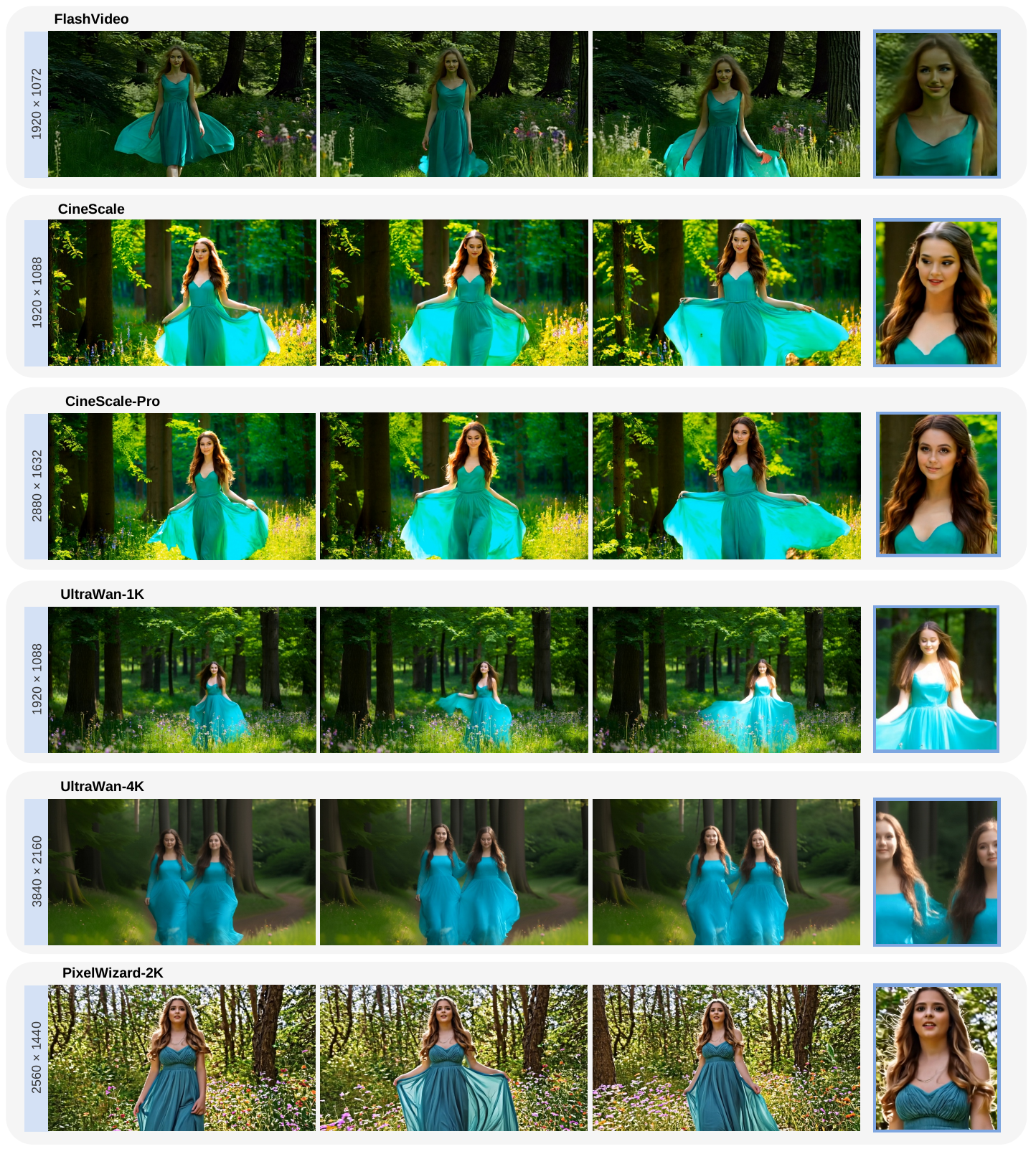}}
\caption{Visual comparison with high-resolution video generation methods. }
\label{Fig.apd.compare5}
\end{figure}
\begin{figure*}
\centering{\includegraphics[width=0.95\textwidth]{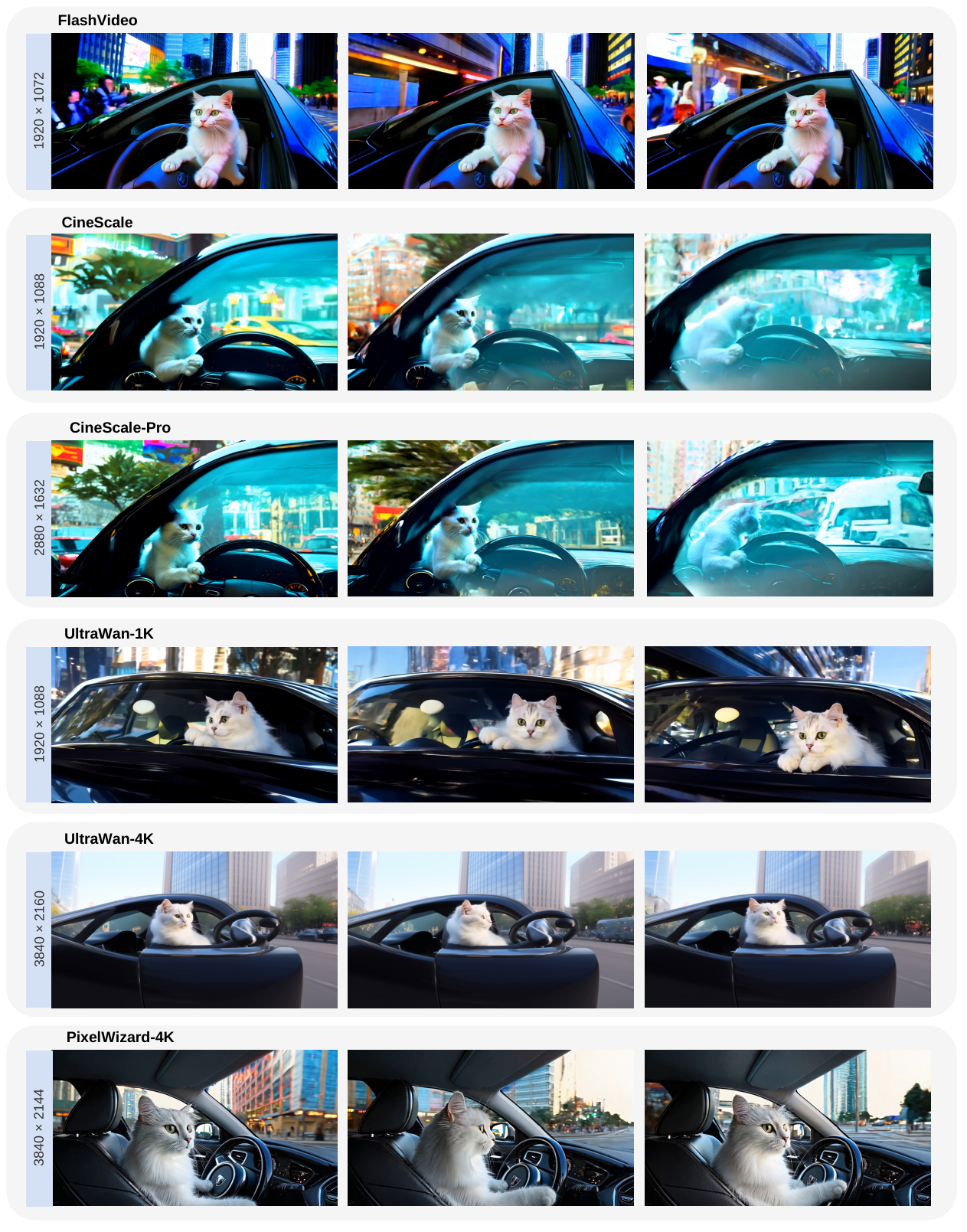}}
\caption{Visual comparison with high-resolution video generation methods. }
\label{Fig.apd.compare2}
\end{figure*}

\begin{figure*}
\centering{\includegraphics[width=0.99\textwidth]{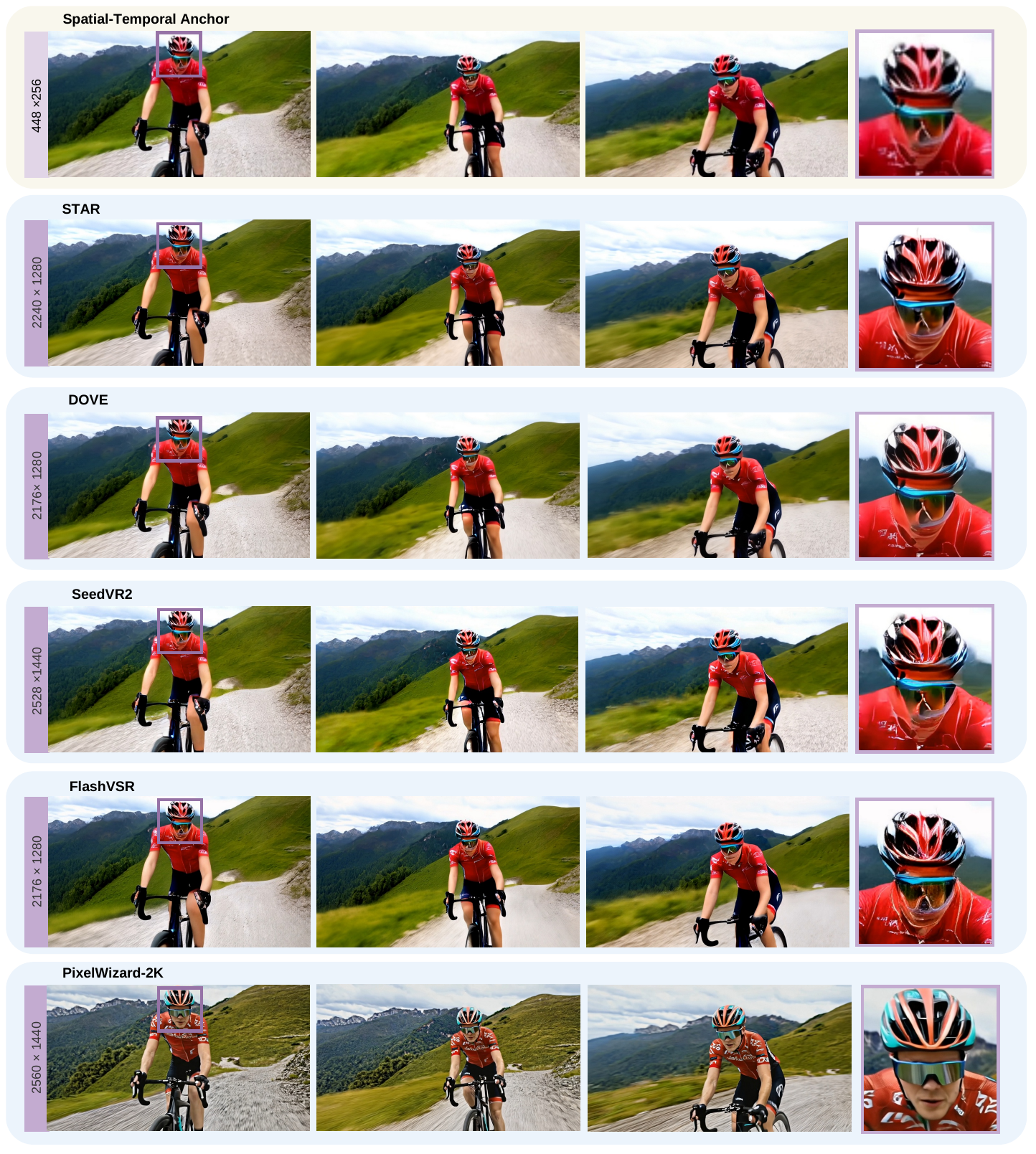}}
\caption{Visual comparison with video SR methods. }
\label{Fig.apd.compare_vsr1}
\end{figure*}
\begin{figure*}
\centering{\includegraphics[width=0.99\textwidth]{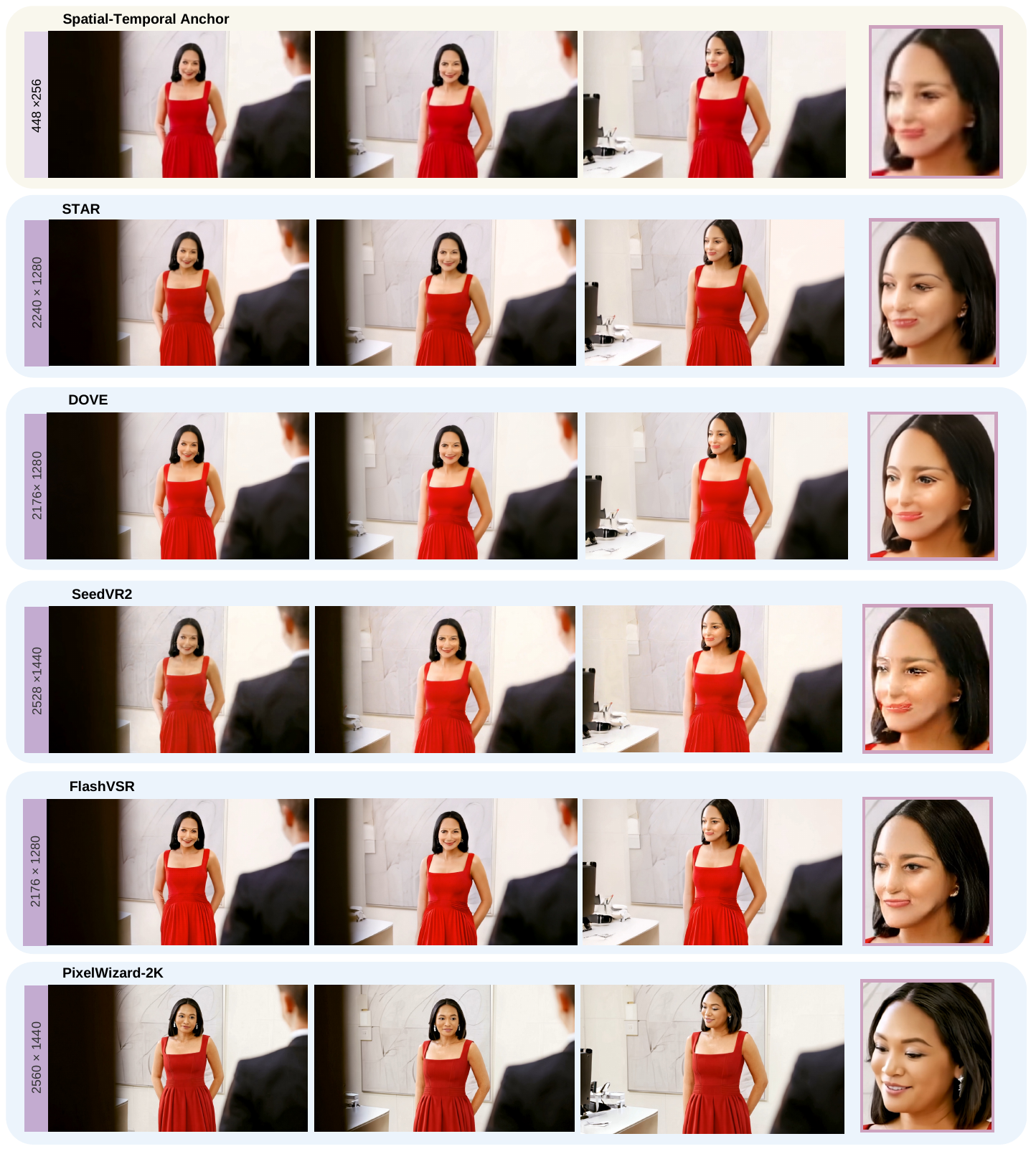}}
\caption{Visual comparison with video SR methods. }
\label{Fig.apd.compare_vsr2}
\end{figure*}

\subsection{More Visual Comparison of Video SR Models}

We present further qualitative comparisons with representative video super-resolution (SR) models in Fig.~\ref{Fig.apd.compare_vsr1},~\ref{Fig.apd.compare_vsr2}.
Our results exhibit more coherent fine-grained details, while avoiding common artifacts such as and structural distortions.

\begin{figure*}
\centering{\includegraphics[width=0.98\textwidth]{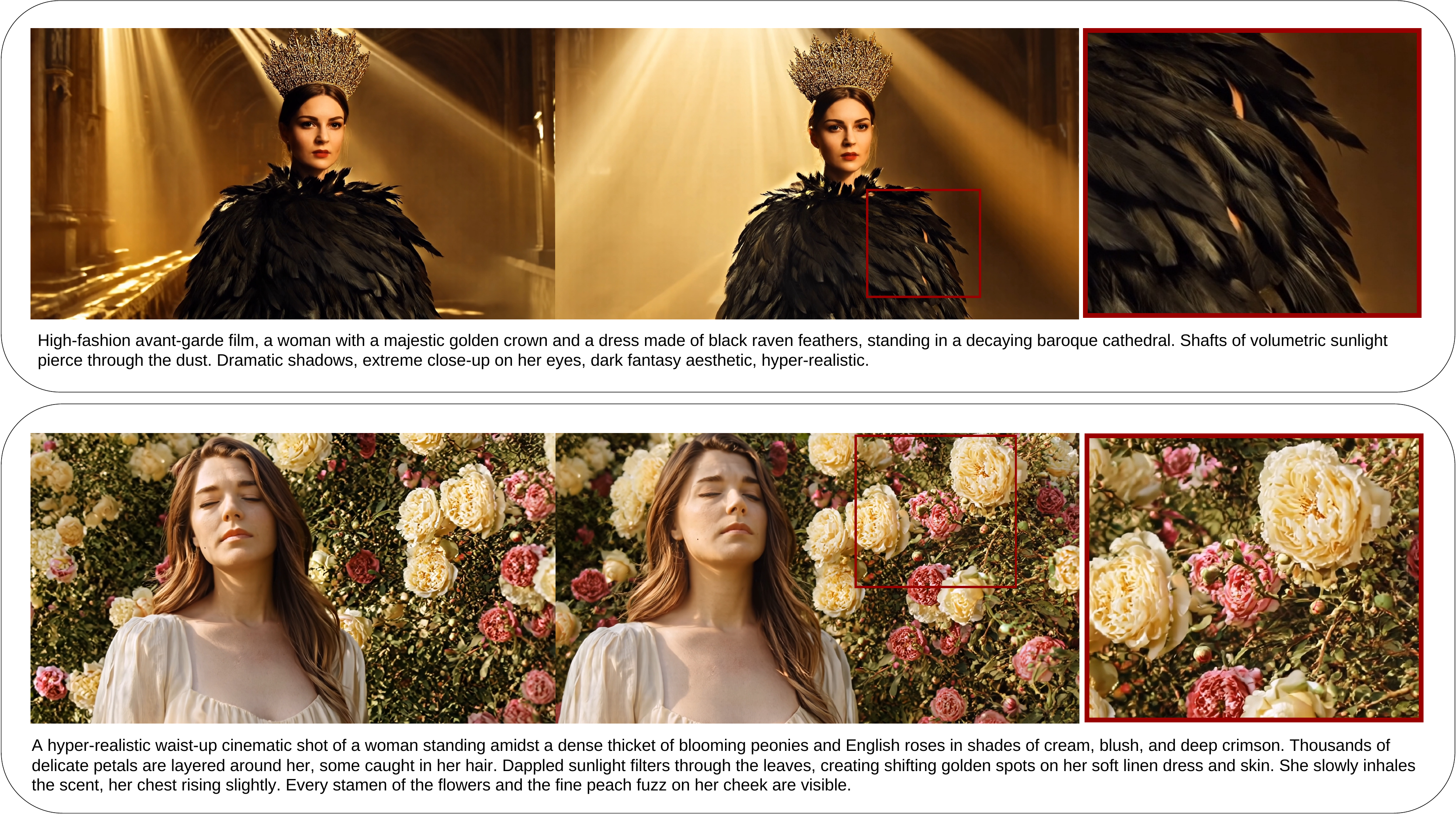}}
\caption{Visualization results of PixelWizard at 2560$\times$1440 resulution.}
\label{Fig.apd.1}
\end{figure*}
\begin{figure*}
\centering{\includegraphics[width=0.98\textwidth]{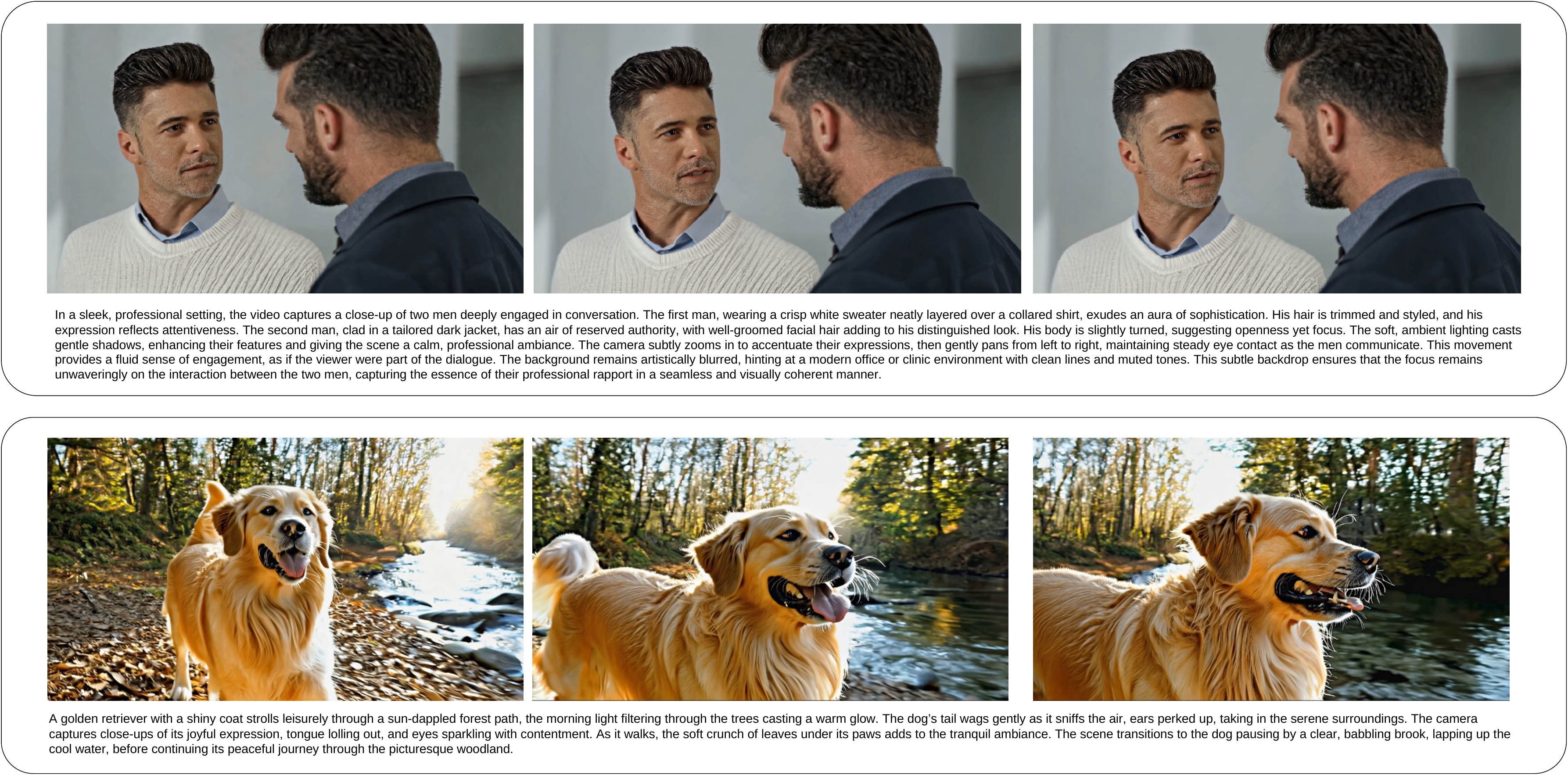}}
\caption{Visualization results of PixelWizard at 2560$\times$1440 resulution.}
\label{Fig.apd.2}
\end{figure*}
\begin{figure*}
\centering{\includegraphics[width=0.9\textwidth]{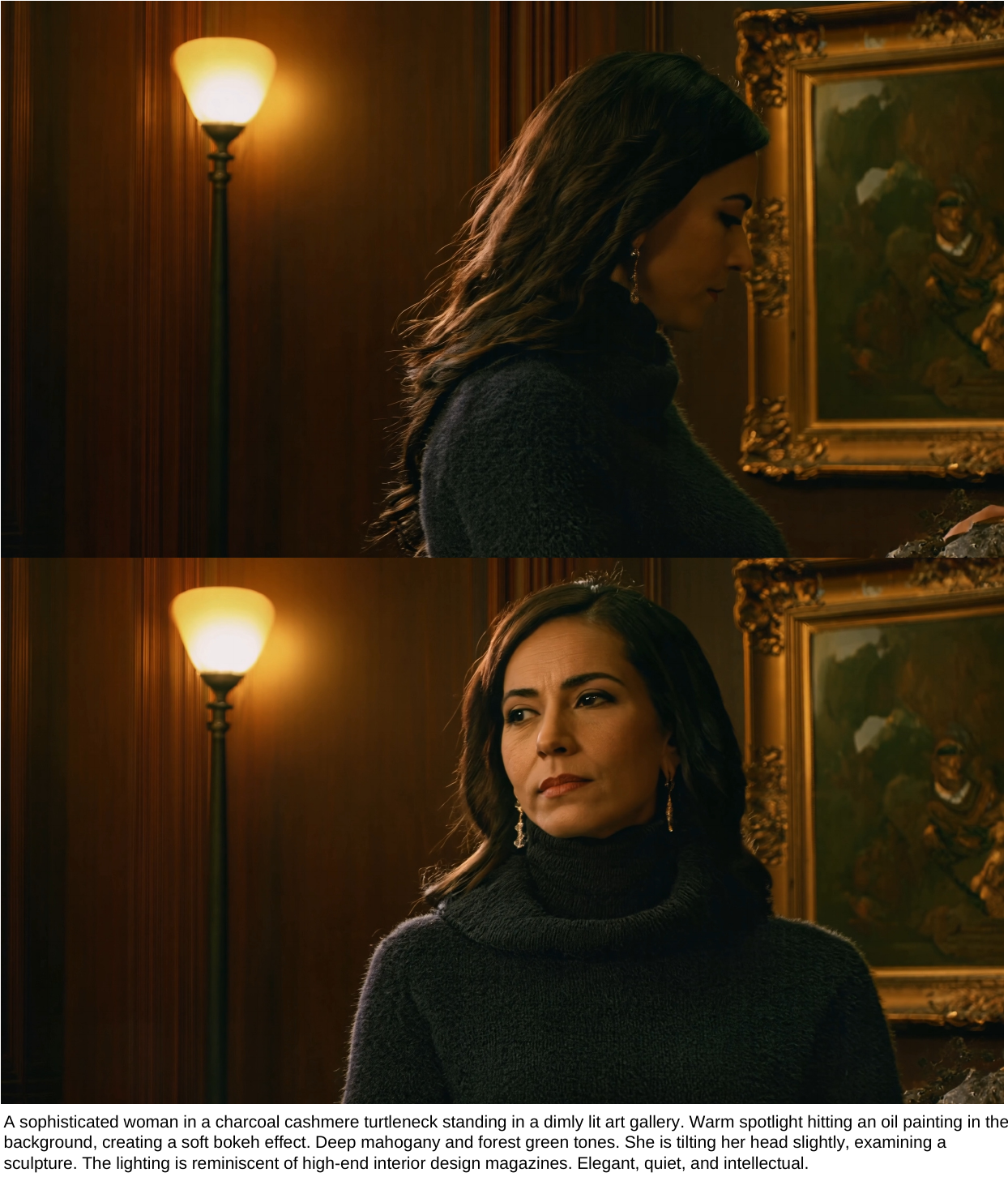}}
\caption{Visualization results of PixelWizard at 3840$\times$2144 resulution.}
\label{Fig.apd.3}
\end{figure*}
\begin{figure*}
\centering{\includegraphics[width=0.9\textwidth]{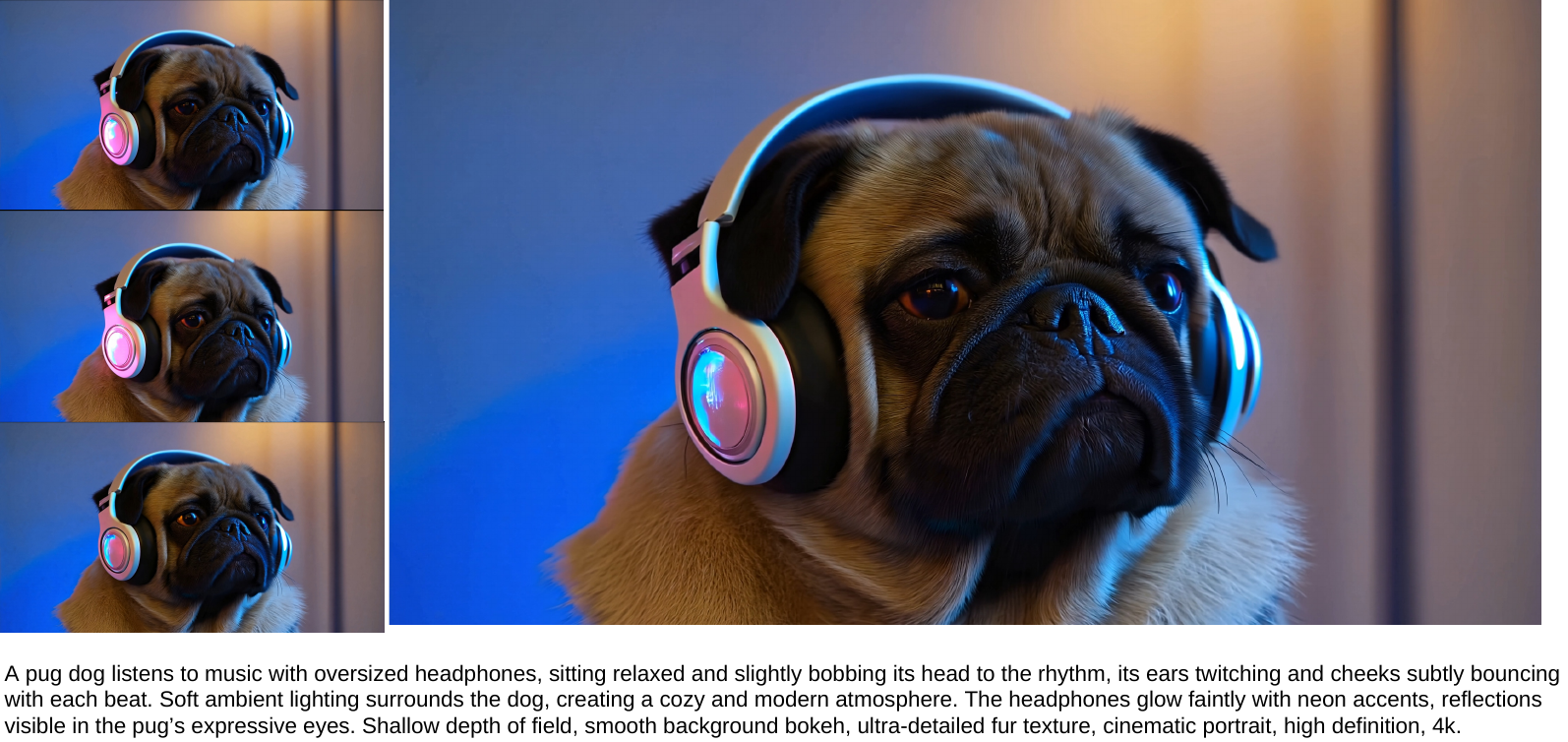}}
\caption{Visualization results of PixelWizard at 3840$\times$2144 resulution.}
\label{Fig.apd.4}
\end{figure*}
\begin{figure*}
\centering{\includegraphics[width=0.9\textwidth]{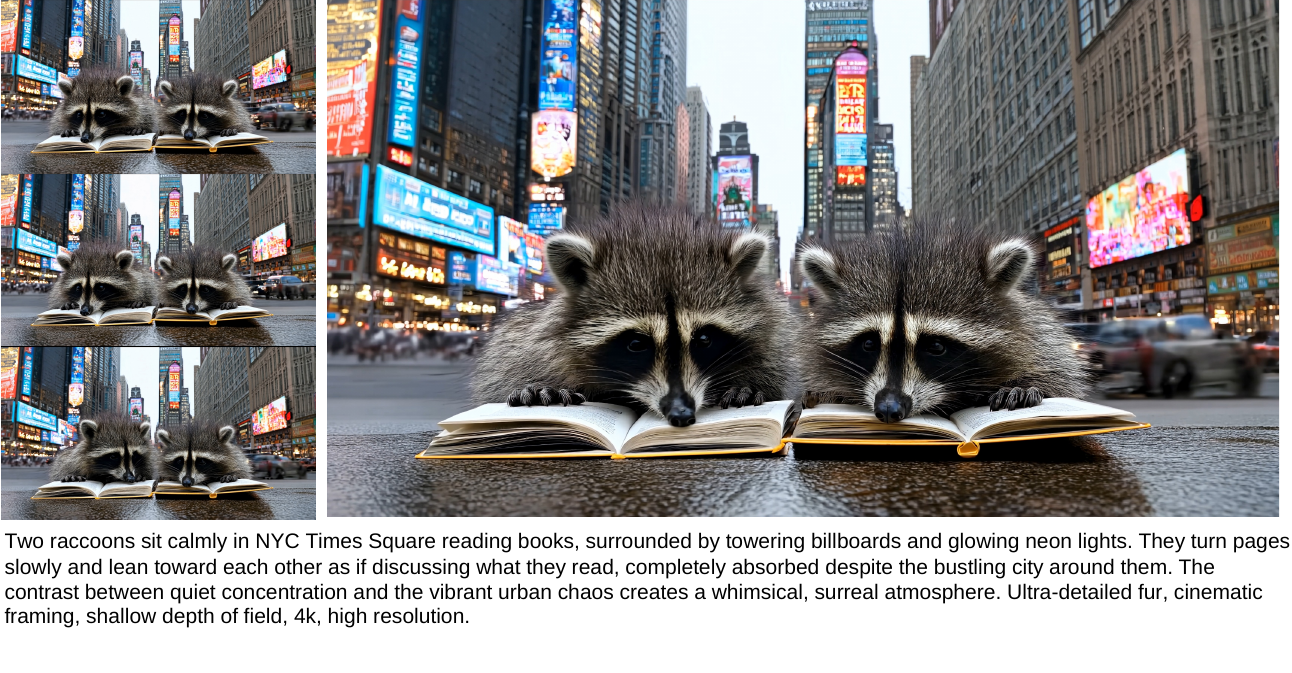}}
\caption{Visualization results of PixelWizard at 3840$\times$2144 resulution.}
\label{Fig.apd.5}
\end{figure*}
\begin{figure*}
\centering{\includegraphics[width=0.9\textwidth]{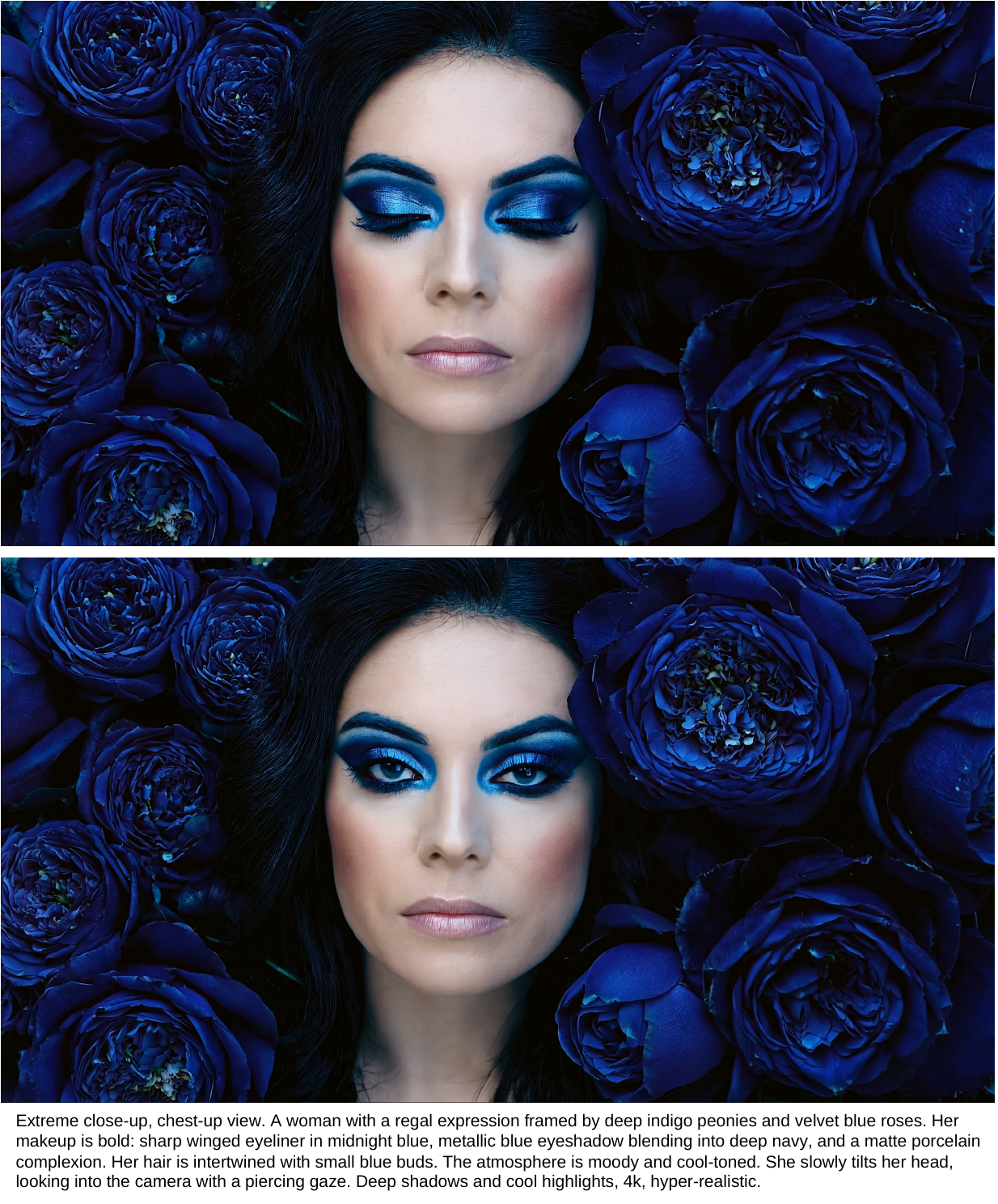}}
\caption{Visualization results of PixelWizard at 3840$\times$2144 resulution.}
\label{Fig.apd.6}
\end{figure*}
\begin{figure*}
\centering{\includegraphics[width=0.9\textwidth]{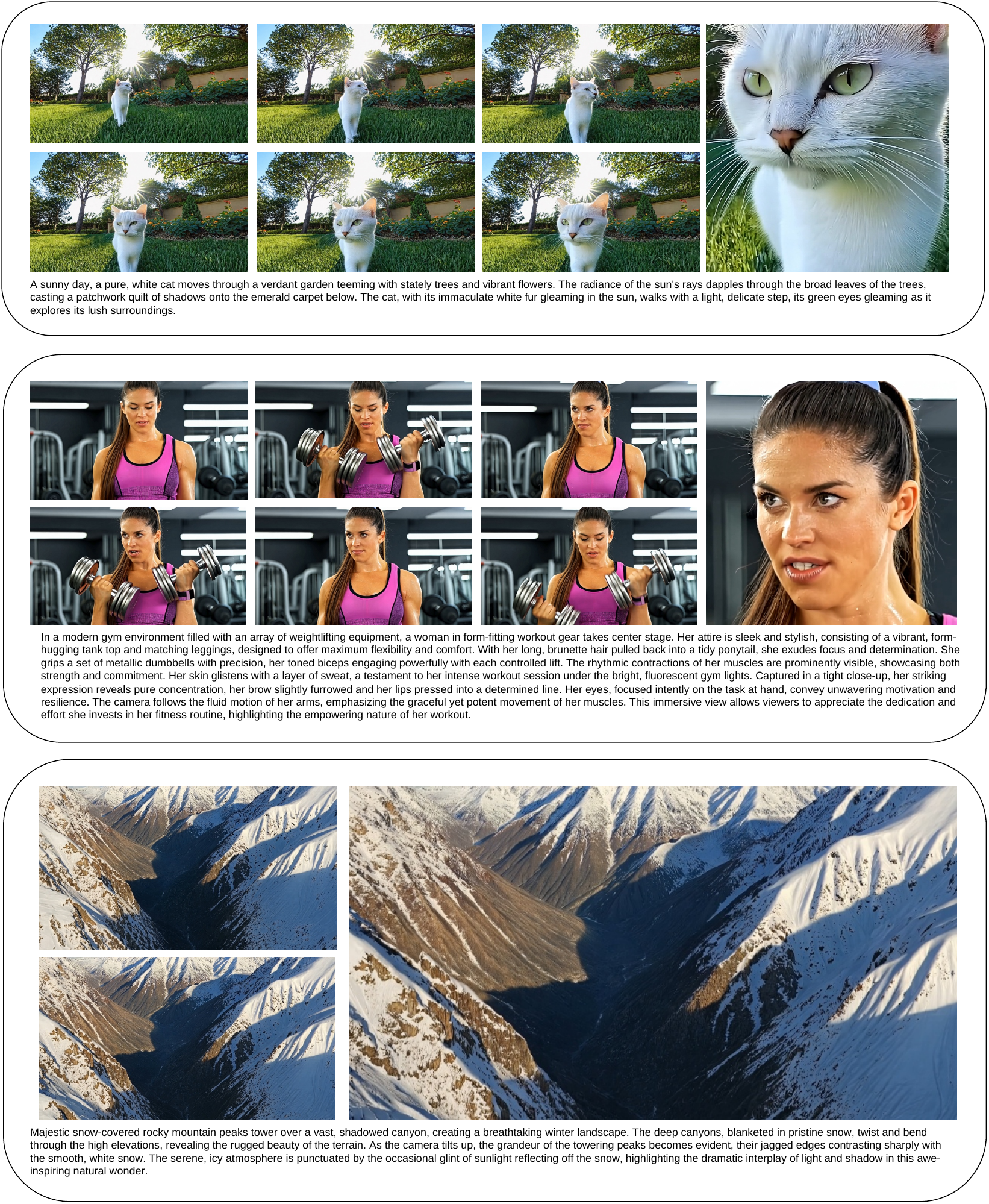}}
\caption{Visualization results of PixelWizard at 3840$\times$2144 resulution.}
\label{Fig.apd.7}
\end{figure*}

\subsection{Additional Visual Results}

We present additional visual results across a wider range of prompts and scenarios.
The results highlight the effectiveness of our method in high-resolution video generation.

\subsection{Text Prompts}

In Table~\ref{Tab.text_prompt}, we provide the text prompts corresponding to the visual examples shown in the figures. 
These prompts are directly used as model inputs and are selected to cover diverse scenes, motions, and visual attributes.

\begin{table*}[t]
\centering
\setlength{\tabcolsep}{1mm}
\small
\caption{Text prompts corresponding to the examples shown in the figures.} 
\label{Tab.text_prompt}
\begin{tabular}{p{0.12\textwidth}p{0.83\textwidth}}
\toprule
\textbf{Figure} & \textbf{Prompt} \\
\midrule

{Figure~\ref{fig:teaser}} & In a grand baroque setting, a regal cat elegantly perches atop an ornate bronze pedestal. Its fiery orange fur is adorned with intricate gold leaf designs, while its emerald eyes assessively survey the room. The cat's fanned tail, adorned with pearls and ribbons, curls gracefully around its slender body. Nearby, a velvet cushion with gold embroidery invites the feline for repose. \\
&  \cellcolor{gray!20} An epic panoramic view from above the clouds, looking down into a hidden mountain valley during peak autumn. The valley floor is an explosion of high-density colors: deep crimson maples, bright yellow larches, and dark evergreen pines, all partially veiled by a moving sea of soft white clouds. Sharp, grey limestone peaks erupt through the cloud layer like islands in an ocean. The lighting is the soft, golden glow of early morning, creating a dreamlike, ethereal masterpiece with infinite layers of depth. \\
& An immense red rock canyon system with thousands of towering sandstone pillars, natural arches, and deep winding gorges. The ground is a chaotic mix of orange sand dunes, scattered desert shrubs, and weathered boulders. A wide emerald river with white-water rapids snakes through the canyon floor, flanked by lush green cottonwood trees and sandy banks. The sky is a fiery explosion of purple and gold sunset clouds, casting long, moving shadows across the complex geological layers. \\
  & \cellcolor{gray!20} A Jack Russell terrier dog snowboards down a steep snowy slope, leaning sharply into the turn as snow sprays dramatically from beneath the board. The dog’s focused expression shows determination and excitement, its ears flapping in the icy wind. Towering alpine mountains rise in the background under a crisp winter sky. Dynamic action shot, motion blur on snow particles, sharp subject focus, cold cinematic lighting.\\
& A dreamy close-up video of a woman’s face and shoulders, framed by oversized glowing blue orchids that seem to pulsate with light. Her makeup features iridescent blue-to-silver gradient eyeshadow and delicate white freckles painted across her nose. The background is a soft blue blur. She blinks slowly, her long lashes casting soft shadows. Hyper-realistic, magical atmosphere.\\
\bottomrule
\end{tabular}
\end{table*}

\section{Limitations and Future Work}
\label{Sec.E}
Spatial–temporal modeling is a key factor that governs how motion, structure, and object interactions evolve over time in generated videos. We observe that certain motion patterns remain challenging to model faithfully. In particular, complex or fast motions may exhibit local deformation, spatial misalignment, or temporally inconsistent trajectories, resulting in behaviors that do not fully conform to real-world dynamics or physical intuition. 
Incorporating stronger motion-aware or physics-informed objectives, for example by aligning motion priors from complementary models or by leveraging reinforcement learning–based optimization, remains an important direction for future work.

Although our framework achieves efficient high-resolution generation, further acceleration remains possible. Existing techniques such as sparse attention and model quantization could be integrated to reduce computational overhead. In addition, the spatial–temporal modeling stage still offers room for speed improvements, for example through distillation-based acceleration to reduce inference steps or model complexity.

\section{License and Usage Statement}
This work uses the UltraVideo dataset from \url{https://huggingface.co/datasets/APRIL-AIGC/UltraVideo}. 
UltraVideo is used solely for academic research purposes in this paper, including model training, evaluation, and reproducibility. 
The authors confirm that the dataset has not been used for any commercial activity.

\end{document}